\documentclass[12pt,peerreview,compsoc,onecolumn]{IEEEtran}

\usepackage{amssymb}

\usepackage{amsmath}
\usepackage{hyperref}
\usepackage{graphicx}
\usepackage{amsmath}

\usepackage{subfig}

\usepackage{graphicx}
\usepackage{amsfonts}
\usepackage{bm}
\usepackage{wrapfig}
\usepackage{makeidx}
\usepackage{moreverb}
\usepackage[lined,boxed,commentsnumbered]{algorithm2e}
\usepackage{multirow}
\usepackage{array}
\usepackage{MnSymbol}
\usepackage{bbding}
\usepackage{booktabs,tabularx}

\DeclareMathOperator*{\argmax}{arg\,max}

\DeclareGraphicsExtensions{.jpg,.png,.tif}
\begin{document}
\title{A New Fuzzy Stacked Generalization Technique and Analysis of its Performance}


\author{Mete~Ozay,~\IEEEmembership{Student Member,~IEEE,}
        Fatos~T.~Yarman~Vural,~\IEEEmembership{Member,~IEEE}
\IEEEcompsocitemizethanks{\IEEEcompsocthanksitem M. Ozay and F. T. Yarman Vural are with the Department of Computer Engineering, Middle East Technical University, Ankara, Turkey.\protect\\
E-mail: {mozay@metu.edu.tr,~vural@ceng.metu.edu.tr}}}

\IEEEtitleabstractindextext{
\begin{abstract}
In this study, a new Stacked Generalization technique called Fuzzy Stacked Generalization (FSG) is proposed to minimize the difference between $N$-sample and large-sample classification error of the Nearest Neighbor classifier. The proposed FSG employs a new \textit{hierarchical} distance learning strategy to minimize the error difference. For this purpose, we first construct an ensemble of base-layer fuzzy $k$- Nearest Neighbor ($k$-NN) classifiers, each of which receives a different feature set extracted from the same sample set. The fuzzy membership values computed at the \textit{decision space} of each fuzzy $k$-NN classifier are concatenated to form the feature vectors of a \textit{fusion space}. Finally, the feature vectors are fed to a meta-layer classifier to learn the degree of accuracy of the decisions of the base-layer classifiers for meta-layer classification.  

The proposed method is examined on both artificial and real-world benchmark datasets. Experimental results obtained using artificial datasets show that the classification performance of the FSG depends on how the individual classifiers share feature vectors of samples. Rather than the power of the individual base layer-classifiers, diversity and cooperation of the classifiers become an important issue to improve the overall performance of the proposed FSG. A weak base-layer classifier may boost the overall performance more than a strong classifier, if it is capable of recognizing the samples, which are not recognized by the rest of the classifiers, in its own feature space. The experiments explore the type of the collaboration among the individual classifiers required for an improved performance of the suggested architecture. Experiments on multiple feature real-world datasets show that the proposed FSG performs better than the state of the art ensemble learning algorithms such as Adaboost, Random Subspace and Rotation Forest. On the other hand, compatible performances are observed in the experiments on single feature multi-attribute datasets. 
\end{abstract}

}

\maketitle
\section{Introduction}
\label{sec:intro}

Stacked Generalization algorithm, proposed by Wolpert \cite{Wolpert} and used by many others \cite{Ueda,sen,Rooney,Zenko,Tan,Ting}, is a widely used ensemble learning technique. The basic idea is to ensemble several classifiers in a variety of ways so that the performance of the Stacked Generalization (SG) is higher than that of the individual classifiers which take place under the ensemble. Although gathering the classifiers under the Stacked Generalization algorithm significantly boosts the performance in some application domains, it is observed that the performance of the overall system may get worse than that of the individual classifiers in some other cases. Wolpert defines the problem of describing the relation between the performance and various parameters of the algorithm as a \textit{black art} problem \cite{Wolpert,Ting}. 

In this study, we suggest a Fuzzy Stacked Generalization (FSG) technique and resolve the \textit{black art} problem \cite{Wolpert} for the minimization of classification error of the nearest neighbor rule. The proposed technique aggregates the independent decisions of the fuzzy base-layer nearest neighbor classifiers by concatenating the membership values of each sample for each class under the same vector space, called the \textit{decision space}. A meta-layer fuzzy classifier is, then, trained to learn the degree of the correctness of the base-layer classifiers. 

There are three major contributions of this study:
\begin{enumerate}

\item We propose a novel \textit{hierarchical distance learning} approach to minimize the difference between $N$-sample and large-sample classification error of the nearest neighbor rule using FSG. The proposed approach enables us to define a  ``distance learning in feature space problem'' as a ``\textit{decision space} design problem'', which is resolved using an ensemble of nearest neighbor classifiers.

\item The proposed FSG algorithm enables us to extract information from different feature spaces using \textit{expert} base-layer classifiers. Expertise of a base-layer classifier on a feature space is analyzed using class membership vectors that reside in a \textit{decision space} of the classifier. Therefore, expertise of base-layer classifiers is used for designing distance functions of the nearest neighbor classifiers. In addition, a \textit{fusion space} of a meta-layer classifier is constructed by aggregating decision spaces of base-layer classifiers. Therefore, the dimension of the feature vectors in the fusion space is fixed to $JC$ where $J$ and $C$ is the number of classifiers and classes, respectively. Then, computational complexity of the meta-layer classifier is $O(NJC)$, where $N$ is the number of samples in the training dataset.

\item We make a thorough empirical analysis of the \textit{black art} problem of the suggested FSG. The empirical results show the effect of the samples which cannot be correctly classified by any of the base-layer classifiers on the classification performance of the FSG. It is observed that if the base-layer classifiers share all the samples in the training set to correctly classify them, then the performance of the overall FSG becomes higher than that of the individual base-layer classifiers. On the other hand, if a sample is misclassified by all of the base-layer classifiers, then this sample causes the performance decrease of the overall FSG.

\end{enumerate}

The suggested Fuzzy Stacked Generalization algorithm is tested on artificial and real datasets by the comparisons with the state of the art ensemble learning algorithms such as Adaboost \cite{adaboost}, Random Subspace \cite{random_subspace} and Rotation Forest \cite{rf}.

In the next section, a literature review of SG architectures and distance learning methods which minimize the generalization error of the nearest neighbor rule is given. The difference between $N$-sample and large-sample errors of the nearest neighbor rule is defined in Section \ref{sec:error_difference}. A distance learning approach which minimizes the error difference is given in Section \ref{sec:dist_learning_error}. Employment of distance learning approach in hierarchical FSG technique and its algorithmic description is given in Section \ref{sec:fsg}. Section \ref{sec:remarks} addresses the conceptual and algorithmic properties of the FSG. Experimental analyses are given in Section \ref{chapter:exp}. Section \ref{sec:fsg_summary} summarizes and concludes the paper.


\section{Related Works and Motivation}
\label{sec:related1}
Various Stacked Generalization architectures are proposed in the literature \cite{Wolpert,Ueda,sen,Rooney,Zenko,Tan,Ghorbani,zhao,iciar,emre,sigletos,ozay_icip,Dzeroski}. Most of them aggregate the decisions of the base-layer classifiers by using vector concatenation operation \cite{Wolpert,Ueda,sen,Rooney,Zenko,Tan,Ghorbani,zhao,iciar,emre,ozay_icip,Dzeroski,wolf,wolf2,Zhouyu},  or majority voting \cite{sigletos} techniques at the meta-layer. 

Ueda \cite{Ueda} employs vector concatenation operation to the feature vectors at the output feature spaces of the base-layer classifiers (which are called decision spaces in our work) and considers these operations as the linear decision combination methods. Then, he compares linear combination and voting methods in an SG, experimentally, where Neural Networks are implemented as the base-layer classifiers. Following the same formulation, Sen and Erdogan \cite{sen} analyze various weighted and sparse linear combination methods by combining the decisions of heterogeneous base-layer classifiers such as decision trees and $k$-NN. Rooney et al. \cite{Rooney} employ homogeneous and heterogeneous classifier ensembles for stacked regression using linear combination rules. Zenko et al. \cite{Zenko} compare the classification performances of SG algorithms, which employ linear combination rules with the other combination methods (e.g. voting) and ensemble learning algorithms (e.g. Bagging and Adaboost). Akbas and Yarman Vural \cite{emre} employed an SG algorithm using fuzzy $k$-NN classifiers for image annotation. Sigletos et al. \cite{sigletos} compare the classification performances of several SG algorithms which combine nominal (i.e., crisp decision values such as class labels) or probabilistic decisions (i.e., estimations of probability distributions). Ozay and Yarman Vural \cite{iciar,ozay_icip} compared the classification performances of the homogeneous base-layer fuzzy $k$-NN classifiers and a linear meta-layer classifier using heterogeneous SG architectures. 

In most of the experimental results given in the aforementioned studies, linear decision combination or aggregation method provides comparable or better performances than the other combination methods. However, performance evaluations of the stacked generalization methods reported in the literature are not consistent with each other. This fact is demonstrated by Dzeroski and Zenko in \cite{Dzeroski} where they employ heterogeneous base-layer classifiers in their stacked generalization architecture. They report that their results contradict with the observations of the studies in the literature on SG. The contradictory results can be attributed to many non-linear relations among the parameters of the SG, such as the number and the structure of base-layer and meta-layer classifiers, and their feature, decision and fusion spaces. 

Selection of the parameters of the SG, and designing classifiers and feature spaces have been considered as a \textit{black art} by Wolpert \cite{Wolpert} and Ting and Witten \cite{Ting}. For instance, popular classifiers, such as, $k$-NN, Neural Networks and Naïve Bayes, can be used as the base-layer classifiers in SG to obtain nominal decisions. However, there are crucial differences among these classifiers in terms of processing the feature vectors. Firstly, $k$-NN and Neural Networks are non-parametric classifiers, whereas the Naïve Bayes is a parametric one. Secondly, $k$-NN is a local classifier which employs the neighborhood information of the features, whereas Neural Networks compute a global linear decision function and Naïve Bayes computes the overall statistical properties of the datasets. Therefore, tracing the feature mappings from base-layer input feature spaces to meta-layer input feature spaces (i.e. fusion spaces) in SG becomes an intractable and uncontrollable problem. Additionally, the outputs of the heterogeneous classifiers give different type of information about the decisions of the classifiers, such as crisp, fuzzy or probabilistic class labeling. 

The employment of fuzzy decisions in the ensemble learning algorithms is analyzed in \cite{Tan,Cho,fvn}. Tan et al. \cite{Tan} use fuzzy $k$-NN algorithms as base-layer classifiers, and employ a linearly weighted voting method to combine the fuzzy decisions for Face Recognition. Cho and Kim \cite{Cho} combine the decisions of Neural Networks which are implemented in the base-layer classifiers using a fuzzy combination rule called fuzzy integral. Kuncheva \cite{fvn} experimentally compares various fuzzy and crisp combination methods, including fuzzy integral and voting, to boost the classifier performances in Adaboost. In their experimental results, the classification algorithms that implement fuzzy rules outperform the algorithms that implement crisp rules. However, the effect of the employment of fuzzy rules to the classification performance of SG is given as an open problem. 

In this study, most of the above mentioned intractable problems are avoided by employing a homogeneous architecture which consists of the same type of base-layer and meta-layer classifiers in a new stacked generalization architecture called Fuzzy Stacked Generalization (FSG). This architecture allows us to concatenate the output decision spaces of the base-layer classifiers, which represent consistent information about the samples. Furthermore, we model linear combination or feature space aggregation method as a feature space mapping from the base-layer output feature space (i.e. decision space) to the meta-layer input feature space (i.e. fusion space). In our proposed FSG, classification rules of base-layer classifiers are considered as the feature mappings from classifier input feature spaces to output decision spaces. In order to control these mappings for tracing the transformations of the feature vectors of samples through the layers of the architecture, homogeneous fuzzy $k$-NN classifiers are used and the behavior of fuzzy decision rules is investigated in both the base-layer and the meta-layer. Moreover, employment of fuzzy $k$-NN classifiers enables us to obtain information about the uncertainty of the classifier decisions and the belongingness of the samples to classes \cite{11,12}.

We analyze the classification error of a nearest neighbor classifier in two parts, namely $i$) $N$-sample error which is the error of a classifier employed on a training dataset of $N$ samples and $ii$) large-sample error which is the error of a classifier employed on a training dataset of large number of samples such that $N \to \infty$. A distance learning approach proposed by Short and Fukunaga \cite{short} is used in a hierarchical FSG architecture from Decision Fusion perspective for the minimization of the error difference between $N$-sample and large-sample error. In the literature, distance learning methods have been employed using prototype \cite{march,march2,li,garc} and feature selection \cite{fw} or weighting  \cite{derrac} methods by computing the weights associated to samples and feature vectors, respectively. The computed weights are used to linearly transform feature spaces of classifiers to more discriminative feature spaces \cite{nca,lmnn,Shalev} in order to decrease large-sample classification error of the classifiers \cite{Paredes}. A detailed literature review of prototype selection and distance learning methods for nearest neighbor classification is given in \cite{garc}.

There are three main differences between our proposed hierarchical distance learning method and the methods introduced in the literature \cite{march,march2,li,garc,fw,derrac,nca,lmnn,Shalev,Paredes}:
\begin{enumerate}
\item The proposed method is used for the minimization of the error difference between $N$-sample and large-sample error, while the aforementioned methods \cite{march,march2,li,garc,fw,derrac,nca,lmnn,Shalev,Paredes} consider the minimization of large-sample error. 
\item We employ a generative feature space mapping by computing the class posterior probabilities of the samples in the decision spaces and use posterior probability vectors as feature vectors in fusion spaces. On the other, the methods given in the literature \cite{march,march2,li,garc,fw,derrac,nca,lmnn,Shalev,Paredes} use discriminative approaches by just transforming the input feature spaces to more discriminative input feature spaces.

\item The aforementioned methods, including the method of Short and Fukunaga \cite{short}, employ distance learning methods in a single classifier. On the other hand, we employ a hierarchical ensemble learning approach for distance learning. Therefore, different feature space mappings can be employed in different classifiers in the ensemble, which enables us more control on the feature space transformations than a single feature transformation in a single classifier.

\end{enumerate}

In Section \ref{sec:error_difference}, we define the problem of minimizing the error difference between $N$-sample and large-sample error in a single classifier. Then, we introduce the distance learning approach for an ensemble of classifiers considering the distance learning problem as a decision space design problem in Section \ref{sec:dist_learning_error}. Employment of the proposed hierarchical distance learning approach in the FSG and its algorithmic description is given in Section \ref{sec:fsg}. We discuss expertise of base-layer classifiers and the dimensionality problems of the feature spaces in FSG, and its computational complexity in Section \ref{sec:remarks}. In order to compare the proposed FSG with the state of the art ensemble learning algorithms, we have implemented Adaboost, Random Subspace and Rotation Forest in the experimental analysis in Section \ref{chapter:exp}. Moreover, we have used the same multi-attribute benchmark datasets with the same data splitting given in \cite{march,march2} to compare the performance of the proposed hierarchical distance learning approach with that of the aforementioned distance learning methods. Since the classification performances of these distance learning methods are analyzed in \cite{march,march2} in detail, we do not reproduce these results in Section \ref{chapter:exp} and refer the reader to \cite{march,march2}.

\section{$N$-sample and Large-sample Classification Errors of $k$-NN} 
\label{sec:error_difference}

Suppose that a training dataset $S=\{ (s_{i} ,y_{i} )\} _{i=1}^{N} $ of $N$ samples, where $y_{i} \in \{ \omega_c \}_{c=1}^C$ is the label of a sample $s_{i} $, is given. A sample $s_i$ is represented in a feature space $F_j$ by a feature vector $\bar{x}_{ij} \in \mathbb{R}^{D_j}$. 

Let $\{P(\bar{x}_{j} | \omega_c) \}_{c=1} ^C$ be a set of probability densities at a feature vector $\bar{x}_{j}$ of a sample $s$, such that $\bar{x}_{j}$ is observed by a given class label $\omega_c$ according to density $P(\bar{x}_{j} | \omega_c)$. Therefore, $P(\bar{x}_{j} | \omega_c)$ is called the likelihood of observing $\bar{x}_{j}$ for a given $\omega_c$. A set of functions $\{ P(\omega_c) \} _{c=1} ^C$ is called the set of prior probabilities of class labels such that $\sum \limits _{c=1} ^C P(\omega_c)=1$ and $P(\omega_c) \geq 0$, $\forall c=1,2, \ldots, C$. Then, the posterior probability of assigning the sample $s$ to a class $\omega_c$ in $F_j$ is computed using the Bayes Theorem \cite{duda} as
\[
P(\omega_c|\bar{x}_{j})=\frac{P(\bar{x}_{j} | \omega_c)P(\omega_c)}{\sum \limits _{c=1} ^C P(\bar{x}_{j} | \omega_c)P(\omega_c)}.
\]
Bayes classification rule estimates the class label $\hat{y }$ of $s$ as \cite{duda}
\[
\hat{y } = \argmax _{\omega_c} \{ P(\omega_c|\bar{x}_{j}) \} _{c=1}^C.
\]

If a loss $\mathcal{L}(\omega_c,s)$ occurs when a sample $s$ is assigned to $\hat{y }={\omega_c }$, then the classification error of the Bayes classifier employed on $F_j$ is defined as \cite{cover}
\[
err(s)=\min _{c}   \sum _{c=1} ^C \mathcal{L}({\omega_c },s) P(\omega_c|\bar{x}_{j})
\]
and the expected error is defined as \cite{cover}
\[
e*= E \{ err(s)  \},
\]
where the expectation is taken over the density $p(\bar{x}_{j})$ of the feature vectors in $F_j$.

In this work, we focus on the minimization of the classification error of a well-known classifier which is $k$ Nearest Neighbors ($k$-NN) \cite{duda}. Given a new test sample $(s',y')$ with $\bar{x}'_{j} \in F_j$, let $\aleph_k(\bar{x}'_{j}) = \{ \bar{x}_{l(1)j}, \ldots, \bar{x}_{l(k)j}\}$ be a set of $k$ nearest neighbors of $\bar{x}'_{j}$ such that 
\[
d(\bar{x}'_{j},\bar{x}_{l(1)j}) \leq d(\bar{x}'_{j},\bar{x}_{l(2)j}) \leq \ldots \leq d(\bar{x}'_{j},\bar{x}_{l(k)j}).
\]
The nearest neighbor rule (e.g. $k=1$) simply estimates $\hat{y'}$, which is the label of $\bar{x}'_{j}$, as the label $y_{l(1)}$ of the nearest neighbor $\bar{x}_{l(1)j}$ of $\bar{x}'_{j}$. In the $k$ nearest neighbor rule (e.g. $k$-NN), $\hat{y'}$ is estimated as
\[
\hat{y'}=\argmax_{ \omega_c } \; \mathcal{N}(\aleph_k(\bar{x}'_{j}), \omega_c),
\]
where $\mathcal{N}(\aleph_k(\bar{x}'_{j}), \omega_c)$ is the number of samples which belong to $\omega_c$ in $\aleph_k(\bar{x}'_{j})$.

Then, the probability of error $\epsilon(\bar{x}_{i,j}, \bar{x}'_{j})=P_N (error|\bar{x}_{i,j}, \bar{x}'_{j})$ of the nearest neighbor rule is computed using $N$ number of samples as
\begin{equation}
\epsilon(\bar{x}_{i,j}, \bar{x}'_{j})=1 - \sum _{c=1} ^C \mu _{c} (\bar{x}_{i,j}) \mu _{c} (\bar{x}'_{j}),
\label{eq:p_error}
\end{equation}
where $\mu _{c} (\bar{x}_{i,j}) = P(\omega_c| \bar{x}_{i,j})$ and $\mu _{c} (\bar{x}'_{j}) = P(\omega_c| \bar{x}'_{j})$ represent posterior probabilities \cite{duda}. 

In the asymptotic of large number of training samples, if $\mu _{c} (\bar{x}'_{j})$ is not singular, i.e. continuous at $\bar{x}'_{j}$, then large-sample error $\epsilon(\bar{x}'_{j}) = \lim \limits_{N \to \infty} P_N (error|\bar{x}'_{j})$ is computed as 
\begin{equation}
\epsilon(\bar{x}'_{j})=1 - \sum _{c=1} ^C \mu ^2 _{c} (\bar{x}'_{j}).
\label{eq:asymp_error}
\end{equation}

It is well known that there is an elegant relationship between the classification errors of Bayes classifier and $k$-NN as follows \cite{cover}:
\[
e* \leq \epsilon(\bar{x}'_{j}) \leq \epsilon(\bar{x}_{i,j}, \bar{x}'_{j}) \leq 2e*.
\]

Then, the difference between the $N$-sample error \eqref{eq:p_error} and the large-sample error \eqref{eq:asymp_error} is computed as
\begin{equation}
\epsilon(\bar{x}_{i,j}, \bar{x}'_{j})-\epsilon(\bar{x}'_{j})=\sum _{c=1} ^C (\mu _{c} (\bar{x}'_{j}))(\mu _{c} (\bar{x}_{i,j})-\mu _{c} (\bar{x}'_{j}) ).
\label{eq:diff_error}
\end{equation}

The main goal of this paper is to minimize the difference between $\epsilon(\bar{x}_{i,j}, \bar{x}'_{j})$ and $\epsilon(\bar{x}'_{j})$ \eqref{eq:diff_error} by employing a distance learning approach suggested by Short and Fukunaga \cite{short} using Fuzzy Stacked Generalization. The distance learning approach of Short and Fukunaga and its employment using a hierarchical distance learning strategy is given in Section \ref{sec:dist_learning_error}. This strategy has been used for modeling the Fuzzy Stacked Generalization and its algorithmic definition is given in Section \ref{sec:fsg}.

\section{Minimization of $N$-sample and Large-sample Classification Error Difference using Hierarchical Distance Learning in the FSG}
\label{sec:dist_learning_error}

Let us start by defining 
\[
e_c(\bar{x}_{i,j},\bar{x}'_{j}) = \big( \mu _{c}(\bar{x}_{i,j})-\mu _{c} (\bar{x}'_{j}) \big ) ^2
\]
and an error function 
\[ 
e(\bar{x}_{i,j},\bar{x}'_{j}) = \sum _{c=1} ^C \epsilon_c(\bar{x}_{i,j},\bar{x}'_{j})
\]
for a fixed test sample $\bar{x}'_{j}$. Then, the minimization of the expected value of the error difference in \eqref{eq:diff_error}, $E _N \{ \big ( \epsilon(\bar{x}_{i,j}, \bar{x}'_{j})-\epsilon(\bar{x}'_{j}) \big )^2 \}$, is equivalent to the minimization of the expected value of the error function \cite{short}
\begin{equation} 
E_N \{ e ^2 (\bar{x}_{i,j},\bar{x}'_{j}) \} ,
\label{eq:exp_error}
\end{equation}
where the expectation is computed over the number of training samples $N$.

Short and Fukunaga \cite{short} notice that \eqref{eq:exp_error} can be minimized by either increasing $N$ or designing a distance function $d(\bar{x}'_{j}, \cdot )$ which minimizes \eqref{eq:exp_error} in the classifiers. In a classification problem, a proper distance function is computed as \cite{short} 
\begin{equation} 
d(\bar{x}'_{j}, \bar{x}_{i,j} )= \|\bar{\mu }(\bar{x}_{i,j}) - \bar{\mu }(\bar{x}'_{j} ) \| _2 ^2, 
\label{eq:dist_single}
\end{equation} 
where 
\[
\bar{\mu }(\bar{x}_{i,j} ) = \left[\mu _{1} (\bar{x}_{i,j} ), \ldots, \mu _{c} (\bar{x}_{i,j} ), \ldots, \mu _{C} (\bar{x}_{i,j} )\right],
\]
\[
\bar{\mu }(\bar{x}'_{j} ) = \left[\mu _{1} (\bar{x}'_{j} ), \ldots, \mu _{c} (\bar{x}'_{j} ), \ldots, \mu _{C} (\bar{x}'_{j} )\right]
\]
and $\| \cdot \| _2 ^2$ is the squared $\ell _2$ norm, or Euclidean distance. 

In a single classifier, \eqref{eq:dist_single} is computed in $F_j, \forall j$, using local approximations to \textit{posterior} probabilities using training  and test datasets \cite{short}. Moreover, if the $N$-sample error is minimized on each different feature space $F_j, \forall j=1,2, \ldots, J$, then an average error over an ensemble of classifiers $\hat{E}_J \{ E_N \{ e ^2 (\bar{x}_{i,j},\bar{x}'_{j}) \} \}$ which is defined as
\begin{equation} 
\hat{E}_J \{ E_N \{ e ^2 (\bar{x}_{i,j},\bar{x}'_{j}) \} \} = \frac{1}{J} \sum \limits _{j=1} ^J E_N \{ e ^2 (\bar{x}_{i,j},\bar{x}'_{j}) \}
\label{eq:error_ens}
\end{equation}
is minimized by using 
\begin{equation} 
d(\bar{x}', \bar{x}_{i})=\sum _{j=1} ^J \sum _{c=1} ^C ( \mu _{c}(\bar{x}_{i,j})-\mu _{c} (\bar{x}'_{j}) )^2.
\label{eq:dist_ens}
\end{equation} 

In this study, an approach to minimize \eqref{eq:error_ens} using \eqref{eq:dist_ens} is employed in a hierarchical decision fusion algorithm. For this purpose, first \textit{posterior} probabilities $\mu _{c}(\bar{x}_{i,j})$ are estimated using individual $k$-NN classifiers, which are called base-layer classifiers. Then the vectors of probability estimates, $\bar{\mu }(\bar{x}_{i,j} )$ and $\bar{\mu }(\bar{x}'_{j} )$, are concatenated to construct 
\[
\bar{\mu }(\bar{x}_{i} ) = \left[\bar{\mu }(\bar{x}_{i,1} ) \ldots \bar{\mu }(\bar{x}_{i,j} ) \ldots \bar{\mu }(\bar{x}_{i,J} ) \right]
\]
and 
\[
\bar{\mu }(\bar{x}' ) = \left[\bar{\mu }(\bar{x}'_{1} ) \ldots \bar{\mu }(\bar{x}'_{j} ) \ldots \bar{\mu }(\bar{x}'_{J} ) \right] ,
\]
for all training and test samples. Finally, classification is performed using $\bar{\mu }(\bar{x}')$ and $\bar{\mu }(\bar{x}_{i} ), \forall i$, by a $k$-NN classifier, called meta-layer classifier, with 
\begin{equation} 
d(\bar{x}', \bar{x}_{i})= \| \bar{\mu }(\bar{x}_{i}) - \bar{\mu }(\bar{x}') \| _2 ^2.
\label{eq:dist_meta}
\end{equation} 

Note that \eqref{eq:dist_meta} can be used for the minimization of the error difference in a feature space $F=F_1 \times F_2 \times \ldots \times F_J$. If $F=F_j$ for $j \in \{1,2,\ldots,J \}$, then \eqref{eq:dist_meta} is equal to \eqref{eq:dist_single}. Therefore, distance learning problem proposed by Short and Fukunaga \cite{short} is reformulated as a decision fusion problem. Then, the distance learning approach is employed using a hierarchical decision fusion algorithm called Fuzzy Stacked Generalization (FSG) as described in the next section. 

\section{Fuzzy Stacked Generalization}
\label{sec:fsg}

Given a training dataset $S=\{ (s_{i} ,y_{i} )\} _{i=1}^{N}$, each sample $s_{i} $ is represented in $J$ different feature spaces $F_{j}$, $j=1,2, \ldots, J $ by a feature vector $\bar{x}_{i,j} \in \mathbb{R}^{D_j} $ which is extracted by using the $j^{th} $ feature extractor $FE _{j}, \forall j=1,2,\ldots,J$. Therefore, training datasets of base-layer classifiers employed on feature spaces $F_{j}, \forall j=1,2,\ldots,J$ can be represented by $J$ different feature sets, $S_{j} =\{ (\bar{x}_{i,j} ,y_{i} )\} _{i=1}^{N} $. 

At the base-layer, each feature vector extracted from the same sample is fed into an individual fuzzy $k$-NN classifier in order to estimate \textit{posterior} probabilities using the class memberships as
\begin{equation}
\mu _{c} (\bar{x}_{i,j} )=\frac{\sum _{n =1}^{k}y_{l(n)}(\left\| \bar{x}_{i,j} -\bar{x}_{l(n),j} \right\|_2)^{-\frac{2}{\varphi -1} }  }{\sum _{n=1}^{k }(\left\| \bar{x}_{i,j} -\bar{x}_{l(n),j} \right\|_2)^{-\frac{2}{\varphi -1} }  } ,
\label{eq:fuzzy_knn}
\end{equation}
where $y_{l(k)}$ is the label of the $k^{th} $-nearest neighbor of $\bar{x}_{i,j}$ which is $\bar{x}_{l(k),j} $, and $\varphi $ is the fuzzification parameter \cite{fknn}. Each base-layer fuzzy $k$-NN classifier is trained and the membership vectors $\bar{\mu }(\bar{x}_{i,j} )$ of each sample $s_{i} $ is computed using leave-one-out cross validation. For this purpose, \eqref{eq:fuzzy_knn} is employed for each $(\bar{x}_{i,j} ,y_{i} )$ using a validation set $S_{j}^{CV} =S_{j} -(\bar{x}_{i,j} ,y_{i} )$, where $(\bar{x}_{l(k),j} ,y_{l(k)} )\in S_{j}^{CV} $.

The class label of an unknown sample $s_{i}$ is estimated by a base-layer classifier employed on $F_j$ as
\[
\hat{y}_{i,j} =\argmax_{\omega_c} (\bar{\mu }(\bar{x}_{i,j} )).
\]
The training performance of the $j^{th} $ base-layer classifier is computed as,   
\begin{equation} \label{eq:perf_tr} 
Perf_{j}^{tr} =\frac{1}{N} \sum _{i=1}^{N}\delta _{\hat{y}_{i,j} } (S_{j} ) ,            
\end{equation} 
where 
\begin{equation} \label{GrindEQ__3_} 
\delta _{\hat{y}_{i,j} } (S_{j} )=
\begin{cases}
1, & \text{if } y_{i} \equiv \hat{y}_{i,j} \\
0, & \text{otherwise}
\end{cases}
\end{equation} 
is the Kronecker delta which takes the value $1$ when the $j^{th} $ base-layer classifier correctly classifies a sample $s_{i} \in S_j$ such that $y_{i} \equiv \hat{y}_{i,j} $. 

When a set of test samples $S_{j}^{te}=\{ s'_{i} \} _{i=1}^{N'}$ is received, the feature vectors $\{ \bar{x}'_{i,j} \} _{i=1}^{N'}$ of the samples are extracted by each $FE_j$. Then, posterior probability $\mu _{c} (\bar{x}'_{i,j})$ of each test sample $s'_{i}$, $i=1,2, \ldots, N'$ is estimated using the training datasets $S_j$ by each base-layer $k$-NN classifier at each $F_j$, $\forall j=1,2, \ldots, J$. 

If a set of labels of test samples, $\{ y'_{i}\} _{i=1}^{N'}$, is available, then the test performance is computed as 
\[
Perf_{j}^{te} =\frac{1}{N'} \sum _{i=1}^{N'}\delta _{\hat{y}'_{i,j} } (S_{j}^{te} ).
\]

The output space of each base-layer classifier is spanned by the class membership vectors $\mu _{c} (\bar{x}_{i,j} )$ of each sample $s_i$. It should be noted that the class membership vectors satisfy 
\[\sum _{c=1}^{C}\mu _{c} (\bar{x}_{i,j} ) =1   .\] 

This equation aligns each sample on the surface of a simplex at the output space of a base-layer classifier, which is called a \textit{Decision Space} of that classifier. Therefore, base-layer classifiers can be considered as transformations which map the input feature space of any dimension into a point on a simplex in a $C$ (number of classes) dimensional decision space (for $C=2$, the simplex is reduced to a line). 

Class-membership vectors obtained at the output of each classifier are concatenated to construct a feature space called \textit{Fusion Space} for a meta-layer classifier. The fusion space consists of $CJ$ dimensional feature vectors $\bar{\mu }(\bar{x}_{i} )$ and $\bar{\mu}(\bar{x}'_{i} )$ which form the training dataset 
\[
S_{meta} =\{ (\bar{\mu }(\bar{x}_{i} ),y_{i} )\} _{i=1}^{N} 
\]
and the test dataset 
\[
S'_{meta} =\{ \bar{\mu }(\bar{x}'_{i} )\} _{i=1}^{N'} 
\]
for the meta-layer classifier. Note that 
\[
\sum _{j=1}^{J}\sum _{c=1}^{C}\mu _{c} (\bar{x}_{i,j} )=J \; \; {\rm and} 
\; \; \sum _{j=1}^{J}\sum _{c=1}^{C}\mu _{c} (\bar{x}'_{i,j} )=J.
\]

Finally, a meta-layer fuzzy $k$-NN classifier is employed to classify the test samples using their feature vectors in $S'_{meta}$ with the feature vectors of training samples in $S_{meta}$. Meta-layer training and test performance is computed as 
\[
Perf_{meta}^{tr} =\frac{1}{N} \sum _{i=1}^{N}\delta _{\hat{y}_{i,meta} } (S_{meta})
\]
and 
\[
Perf_{meta}^{te} =\frac{1}{N'} \sum _{i=1}^{N'}\delta _{\hat{y}'_{i,meta} } (S'_{meta} ),
\]
respectively. An algorithmic description of the FSG is given in Algorithm \ref{alg:fsg}.

\begin{algorithm}
%
\SetKwFunction{Union}{Union}\SetKwFunction{FindCompress}{FindCompress}
\SetKwInOut{Input}{input}\SetKwInOut{Output}{output}
\SetKwComment{tcc}{/*}{*/}
\Input {Training set $S=\{ (s_{i} ,y_{i} )\} _{i=1}^{N} $, test set $S_{j}^{te}=\{ s'_{i}\} _{i=1}^{N'}$ and $J$ feature extractors $FE_j$, $ \forall j=1,2,\ldots,J$. }
\Output{Predicted class labels of the test samples $\{ \hat{y}'_{i}\} _{i=1}^{N'}$.}
\ForEach{$j=1,2,\ldots,J$}{ 

\nl Extract features $\{ \bar{x}_{i,j} \} _{i=1}^{N}$ and $\{ \bar{x}'_{i,j} \} _{i=1}^{N'}$ using $FE_j$; \\
\nl Compute $\{ \bar{\mu }(\bar{x}_{i,j} )\} _{i=1}^{N}$  and $\{ \bar{\mu }(\bar{x}'_{i,j} )\} _{i=1}^{N'}$ using \eqref{eq:fuzzy_knn}; \\
}

\nl Construct $S_{meta} :=\{ (\bar{\mu }(\bar{x}_{i} ),y_{i} )\} _{i=1}^{N} $ and $S'_{meta} := \{ \bar{\mu }(\bar{x}'_{i} )\} _{i=1}^{N'}$; \\
\nl Employ meta-layer classification using $S_{meta}$ and $S'_{meta}$ to predict $\{ \hat{y}'_{i}\} _{i=1}^{N'}$; \\
\caption{Fuzzy Stacked Generalization.}
\label{alg:fsg}
\end{algorithm}


The proposed algorithm has been analyzed on artificial and benchmark datasets in Section \ref{chapter:exp}. A treatment of the FSG is given in the next section.
\newpage
\section{Remarks On The Performance Of Fuzzy Stacked Generalization}
\label{sec:remarks}
In this section, we discuss the error minimization properties of the FSG, and the relationships between the performance of the FSG and various learning parameters. 

\subsection{Expertise of the Base-layer Classifiers, Feature Space Dimensionality Problem and Performance of the FSG  }
Employing distinct feature extractors for each classifier enables us to split various attributes of the feature spaces, coherently. Therefore, each base-layer classifier gains an expertise to learn a specific property of a sample, and correctly classifies a group of samples belonging to a certain class in the training data. This approach assures the diversity of the classifiers as suggested by Kuncheva \cite{comb_kun} and enables the classifiers to collaborate for learning the classes or groups of samples. It also allows us to optimize the parameters of each individual base-layer classifier independent of the other. 

Formation of the fusion space by concatenating the decision vectors at the output of base-layer classifiers helps us to learn the behavior of each individual classifier to recognize a certain feature of the sample, which may result in substantial improvement in the performance at the meta-layer. However, this postponed concatenation technique increases the dimension of the feature vector to $CJ$. If one deals with a classification problem of high number of classes, which may also require high number of base-layer classifiers with large number of samples for high performance, the dimension of the feature space at the meta-layer becomes large causing again curse of dimensionality. An analysis to show the decrease in performance as the number of classes and the classifiers increase is provided in \cite{iciar}. More detailed experimental results on the change of the classification performances as the number of feature spaces increases are given by comparing FSG on benchmark datasets with state of the art ensemble learning algorithms in Section \ref{chapter:exp}.

Since there are several parameters such as the number of classes, the number of feature extractors, and the mean and variances of distributions of the feature vectors, which affect the performance of classifier ensembles, there is no generalized model that defines the behavior of the performance with respect to these parameters. However, it is desirable to define a framework which ensures an increase in the performance of the FSG compared to the performance of the individual classifiers. 

In addition, the design of the feature spaces of individual base-layer classifiers, size of the training set, number of classes and the relationship between all of these parameters affect the performance. A popular approach to design the feature space of a single classifier is to extract all of the relevant features from each sample, and aggregate them under the same vector. Unfortunately, this approach creates the well-known dimensionality curse problem. On the other hand, reducing the dimension by the methods such as principal component analysis, normalization, and feature selection algorithms may cause the loss of information. Therefore, one needs to find a balance between the dimensionality curse and the information deficiency in designing the feature space. 

The suggested FSG architecture establishes this balance by designing independent base-layer fuzzy $k$-NN classifiers each of which receives relatively low dimensional feature vectors compared to the concatenated feature vectors of the single classifier approach. This approach avoids the problem of normalization required after the concatenation operation. Note that the dimension of the decision space is independent of the dimensions of the feature spaces of the base-layer classifiers. Therefore, no matter how high is the dimension of the individual feature vectors at the base-layer, this architecture fixes the dimensions at the meta-layer to $CJ$ (number of classes $\times$ number of feature extractors). This may be considered as a partial solution to dimensionality curse problem provided that $CJ$ is bounded to a value to assure statistical stability to avoid curse of dimensionality.

\subsection{Computational Complexity of the FSG}
\label{sec:fsg_complexity}

In the analysis of the computational complexities of the proposed FSG algorithm, computational complexities of feature extraction algorithms are ignored assuming that the feature sets are already computed and given. 
 
The computational complexity of the Fuzzy Stacked Generalization algorithm is dominated by the number of samples. The computational complexity of a base-layer $k$-NN classifier is $O(ND_j)$, $\forall j=1, \ldots,J$. If each base-layer classifier is implemented by an individual processor in parallel, then the computational complexity of base-layer classification process is $O(N\tilde{D})$, where $\tilde{D} = \max \{ D_j \}_{j=1}^J$. In addition, the computational complexity of a meta-layer classifier which employs a fuzzy $k$-nn is $O(NJC)$. Therefore, the computational complexity of the FSG is $O(N (\tilde{D} + JC))$. 

In the following section, we provide an empirical study to analyze the remarks given in this section.

\section{Experimental Analysis}
\label{chapter:exp}

In this section, three sets of experiments are performed to analyze the behavior of the suggested FSG and compare its performance with the state of the art ensemble learning algorithms.

\begin{enumerate}
\item In order to examine the proposed algorithm for the minimization of the difference between the $N$-sample and the large-sample classification error, we propose an artificial dataset generation algorithm following the comments of 

\begin{itemize}

\item Cover and Hart \cite{cover} on the analysis of the relationship between the class conditional densities of the datasets and the performance of the nearest neighbor classification algorithm, and
\item Hastie and Tibshirani \cite{hastie} on the development of metric learning methods for $k$-NN.

\end{itemize}

In addition, we analyze the relationship between performances of base-layer and meta-layer classifiers considering sample and feature shareability among base-layer classifiers and feature spaces. Then, we examine geometric properties of transformations between feature spaces by visualizing the feature vectors in the spaces and tracing the samples in each feature space, i.e. base-layer input feature space, base-layer output decision space and meta-layer input fusion space.

\item Next, benchmark pattern classification datasets such as Breast Cancer, Diabetis, Flare Solar, Thyroid, German, Titanic \cite{march,march2,li,garc,w,blake}, Caltech 101 Image Dataset \cite{caltech} and Corel Dataset \cite{iciar} are used to compare the classification performances of the proposed approach and state of the art supervised ensemble learning algorithms. We have used the same data splitting of the benchmark Breast Cancer, Diabetis, Flare Solar, Thyroid, German and Titanic datasets suggested in \cite{march,march2} to enable the reader to compare our results with the aforementioned distance learning methods referring to \cite{march,march2}. 

\item Finally, we examine FSG in a real-world target detection and recognition problem using a multi-modal dataset. The dataset is collected using a video camera and microphone in an indoor environment to detect and recognize two moving targets. The problem is defined as a four-class classification problem, where each class represents absence or presence of the targets in the environment. In addition, we analyze the statistical properties of the feature spaces by computing entropy values of the distributions of the feature vectors in each feature space, and comparing the entropy values of each feature space of each classifier computed at each layer.

\end{enumerate}

In the FSG, $k$ values of the fuzzy $k$-NN classifiers are optimized by searching $k \in \{ 1,2,\ldots,\sqrt{N} \}$ using cross validation, where $N$ is the number of samples in training datasets. In the experiments, fuzzy $k$-NN is implemented both in Matlab\footnote{An Matlab implementation is available on https://github.com/meteozay/fsg.git} and C++. For C++ implementations, a fuzzified modification of a GPU-based parallel $k$-NN is used \cite{gpu_knn}. Classification performances of the proposed algorithms are compared with the state of the art ensemble learning algorithms, such as Adaboost \cite{adaboost}, Random Subspace \cite{random_subspace} and Rotation Forest \cite{rf}. Weighted majority voting is used as the combination rule in Adaboost. Decision trees are implemented as the weak classifiers in both Adaboost and Rotation Forest, and $k$-NN classifier is implemented as the weak classifier in Random Subspace. The number of weak classifiers $Num_{weak} \in \{ 1,2, \ldots, 2D \}$ is selected using cross-validation in the training set, where $D= \sum \limits _{j=1} ^J D_j$ is the dimension of the feature space of the samples in the datasets.  Adaboost and Random Subspace algorithms are implemented using Statistics Toolbox of Matlab. 

Experimental analyses of the proposed FSG algorithm on artificial datasets are given in Section \ref{sec:synthetic}. In Section \ref{sec:bench_fsg}, classification performances of the proposed algorithms and the state-of-the art classification algorithms are compared using benchmark datasets.

\subsection{Experiments on Artificial Datasets}
\label{sec:synthetic}
The relationship between the performance of the $k=1$ and $k \geq 2$ nearest neighbor algorithms and the statistical properties of the datasets has been studied in the last decade by many researchers. Cover and Hart \cite{cover} analyzed this relationship with an elegant example, which is revised later by Devroye, Gyorfi and Lugosi \cite{devroye}. 

In the example, suppose that the feature vectors of the samples of a training dataset $\{ (s_i,y_i) \} _{i=1}^N$ are grouped in two disks with centers $\bar{o}_1$ and $\bar{o}_2$, which represent the class groups $\omega_1$ and $\omega_2$ such that $\parallel \bar{o}_1 - \bar{o}_2 \parallel _2 \geq \sigma ^{1,2} _{BC}$ in a two dimensional feature space, where $\sigma^{1,2}_{BC}$ is the between-class variance. In addition, assume that the class conditional densities are uniform and 
\[
P(\omega_1)=P( \omega_2)=\frac{1}{2}.
\] 

Note that the probability that $n$ samples belong to the first class $\omega_1$, i.e. that the feature vectors reside in the first disk, is 
\[ 
\frac{1}{2^N} { N \choose n}.
\]

Now, assume that the feature vector of a training sample $s_i$ belonging to $\omega_1$ is classified by $k=1$ nearest neighbor rule. Then, $s_i$ will be misclassified if its nearest neighbor resides in the second disk. However, if the nearest neighbor of $s_i$ resides in the second disk, then each of the feature vectors must reside in the second disk. Therefore, the classification error is the probability that all of the samples reside in the second disk such that
\[
P(y_i \in \omega_1, y_{j \neq i} \in \omega_2)+P(y_i \in \omega_2, y_{j \neq i} \in \omega_1)=\frac{1}{2^N}.
\]

If $k$-NN rule is used for classification with $k=2\hat{k}+1$, where $\hat{k} \geq 1$, then an error occurs if $\hat{k}$ or less number of features reside in the first disk with probability
\[
P(y_i \in \omega_1, \sum \limits _{j=1} ^N I( y_{j \neq i} \in \omega_1) \leq \hat{k} )  + P(y_i \in \omega_2, \sum \limits _{j=1} ^N I( y_{j \neq i} \in \omega_2) \leq \hat{k} )
\]
which is a Binomial distribution $Binomial(\hat{k},N,\frac{1}{2})$
\[
\sum \limits _{n=0} ^{\hat{k}}  { N \choose n } \big (\frac{1}{2} \big)^n \big (1-\frac{1}{2} \big)^{N-n} = \big ( \frac{1}{2} \big ) ^N \sum \limits _{n=0} ^{\hat{k}}  { N \choose n }. 
\]
Then the following inequality holds 
\[
\big ( \frac{1}{2} \big ) ^N \sum \limits _{n=0} ^{\hat{k}}  { N \choose n } > \big ( \frac{1}{2} \big)^N.
\]

Therefore, the classification or generalization error of the $k$-NN depends on the class conditional densities \cite{cover} such that $k=1$ rule performs better than $k \geq 2$ rule when the between class variance of the data distributions $\sigma^{c,c'}_{BC}$ is smaller than the within class variances $\Sigma _{c}$, $\forall c \neq c'$, $c=1,2,\ldots,C$, $c' =1,2,\ldots,C$.

Although Cover and Hart \cite{cover} introduced this example to analyze the classification performances of the nearest neighbor rules, Hastie and Tibshirani \cite{hastie} used the results of the example in order to define a metric, which is a function of $\sigma^{c,c'}_{BC}$ and $\Sigma _{c}$, to minimizes the difference between the $N$-sample and large-sample errors. Since the minimization of error difference is one of the motivations of FSG, a similar experimental setup is designed in order to analyze the performance of FSG in this section.

In the experiments, feature vectors of the samples in the datasets are generated using a Gaussian distribution in each $D_j=2$ dimensional feature space $F_j$, $j=1,2,\ldots,J$. While constructing the datasets, the mean vector $\bar{o}_{c} $ and the covariance matrix $\Sigma _{c} $ of  the class-conditional density of a class $\omega_c$ 
\begin{equation}
\label{eq:data_generate}
f(\bar{x}|\; \; {\bar{o}_{c}} ,\; \Sigma _{c} )=\frac{1}{\sqrt{(2\pi )^{d} |\Sigma_{c} |} } \exp \left[-\frac{1}{2} \left(\bar{x}-\bar{o}_{c} \right)^{T} \Sigma _{c} ^{-1} \left(\bar{x}-\bar{o}_{c} \right)\right]
\end{equation} 
are systematically varied in order to observe the effect of the class overlaps to the classification performance. One can easily realize that there are explosive alternatives for changing the parameters of the class-conditional densities in a $D_j$-dimensional vector space. However, it is quite intuitive that the amount of overlaps among the classes affects the performance of the individual classifiers rather than the changes in the class scatter matrix. Therefore, we suffice to control only the amount of overlaps during the experiments. This task is achieved by fixing the covariance matrix $\Sigma _{c}$, in other words within-class-variance, and changing the mean values of the individual classes, which varies the between-class variances, $\sigma ^{c,c'} _{BC} $, $\forall c \neq c' $, $c=1,2,\ldots,C$, $c' =1,2,\ldots,C$.   

Denoting $v _{i} $  as the eigenvector and $\vartheta _{i} $ as the eigenvalue of a covariance matrix $\Sigma $, we have $\Sigma v _{i} =\vartheta _{i} v _{i}$. Therefore, the central position of the sample distribution constructed by datasets in a $2$-dimensional space is defined by $v _{1}$ and $v _{2}$ and the propagation is defined by $\vartheta_{1}^{1/2}$ and $\vartheta_{2}^{1/2}$. In the datasets, covariance matrices are held fix and equal. Therefore, the eigenvalues represented on both axes are the same. As a result, datasets are generated by the circular Gaussian function with fixed radius.

In this set of experiments, a variety of artificial datasets is generated in such a way that \textit{most} of the samples are correctly labeled by at least one base-layer classifier. In other words, feature spaces are generated to construct classifiers which are \textit{expert} on specific classes. The \textit{performances} of the base-layer classifiers are controlled by fixing the covariance matrices, and changing the mean values of Gaussian distributions which are used to generate the feature vectors. Thereby, we can analyze the relationship between classification performance, the number of samples correctly labeled by at least one base-layer classifier and expertise of the base-layer classifiers. 

In order to avoid the misleading information in this gradual overlapping process, the feature vectors of the samples belonging to different classes are first generated apart from each other to assure the linear separability in the initialization step. Then, the distances between the mean values of the distributions are gradually decreased. The ratio of decrease is selected as one tenth of between-class variance of distributions for each class pair $\omega _{c} $ and $\omega _{c' } $, $\forall c \neq c' $, $c=1,2,\ldots,C$, $c' =1,2,\ldots,C,$ which is $\frac{1}{10} \sigma _{BC}^{c,c'} $, where $\sigma _{BC}^{c,c'} =\left\| \bar{o}_{c} -\bar{o}_{c' } \right\| $. The termination condition for the algorithms is
\[
\sum _{c,c' }\sigma _{BC}^{c,c' } =0 ,\forall c \neq c',c=1,2,\ldots,C, c'=1,2,\ldots,C. 
\] 
At each epoch, only the mean value of the distribution of one of the classes approaches to the mean value of that of another class, while keeping the rest of the mean values fixed. Defining $J$ as the number of classifiers fed by $J$ different feature extractors and $C$ as the number of classes, the data generation method is given in Algorithm \ref{alg:data_generation}.

\begin{algorithm}
%
\SetKwFunction{Union}{Union}\SetKwFunction{FindCompress}{FindCompress}
\SetKwInOut{Input}{input}\SetKwInOut{Output}{output}
\Input {The number of feature spaces $J$, the number of classes $C$, the mean value vectors $\bar{o}_c$ and the within class variances $\Sigma_c$ of the class conditional densities, $\forall c=1,2, \ldots, C$. }
\Output{Training and test datasets.}

\ForEach{$j=1,2,\ldots,J$}{
\ForEach{$c'=1,2,\ldots,C$}{
\nl Initialize $\hat{o}_{c'}$; \\
\ForEach{$c=1,2,\ldots,C$}{
\Repeat{$ \sigma_{BC}^{c,c'} \neq 0$}{
\nl Generate feature vectors using \eqref{eq:data_generate};\\ 
\nl $\sigma _{BC}^{c,c' } \leftarrow \left\| \bar{o}_{c} -\hat{o}_{c'} \right\| $ ;\\
\nl $\hat{o}_{c'} \leftarrow \bar{o}_{c} + \frac{1}{10} \sigma _{BC}^{c,c' } $; \\
}}}}
\nl Randomly split the feature vectors into two datasets, namely test and training datasets.
\caption{Artificial data generation algorithm.}
\label{alg:data_generation}
\end{algorithm}

\subsubsection{Performance Analysis on Artificial Datasets}

In the first set of the experiments, $7$ base-layer classifiers are used. The number of samples belonging to each class $\omega_c$ is taken as $250$, and $2$-dimensional feature spaces are fed to each base-layer classifier as input for $C=12$ classes with $250 \times 12=3000$ samples. The feature sets are prepared with fixed and equal 
\[
\Sigma _{c} =\left(\begin{array}{cc} {5} & {5} \\ {5} & {5} \end{array}\right), \forall c =1,2,\ldots,12,
\] 
which is the covariance matrix of the class conditional distributions in $F_{j}$, $\forall j=1,2,\ldots ,7$. In other words, $\vartheta_1 ^{\frac{1}{2}} =5$ and $\vartheta_{2} ^{\frac{1}{2}} =5$. 

The features are distributed with different $\sigma _{BC}^{c,c' }$ and converged towards each other using Algorithm \ref{alg:data_generation}. The matrix $\Omega_{j} =[\bar{o}_{c,j} ]_{c=1}^{12} $, with the row vectors that contain the mean values $\bar{o}_{c,j} $ of the distribution of each class $\omega_{c}$ at each space $j=1,2,\ldots ,7$ is defined  as
\[
\Omega=[\Omega_{1} ,\Omega_{2} ,\Omega_{3} ,\Omega_{4} ,\Omega_{5} ,\Omega_{6} ,\Omega_{7} ].
\] 

In order to analyze the relationship between the number of samples that are correctly classified by at least one of the base-layer classifiers and classification performance of the FSG, the average number of samples that are correctly classified by at least one base-layer classifier, which is denoted as $\hat{Ave}_{corr}$, is also given in the experimental results. 

In each epoch, features belonging to different classes are distributed with different topologies in each classifier by different overlapping ratios. For example, feature vectors of the samples belonging to the ninth class is located apart from that of the rest of the classes in $F_7$, while they are overlapped in other feature spaces. In this way, the classification behaviors of the base-layer classifiers are controlled through the topological distributions of the features, and classification performances are measured by the metrics given in Section \ref{sec:dist_learning_error}.

In Table \ref{tab:fsg_table1}, performances of individual classifiers and the proposed algorithms are given for an instance of the dataset generated by Algorithm \ref{alg:data_generation}, where the datasets are constructed in such a way that each sample is correctly recognized by at least one of the base-layer classifiers, i.e. $\hat{Ave}_{corr}=1$. Although the performances of individual classifiers are in between $53\%-66\%$, the classification performance of FSG is $99.9\%$. In that case, different classes are distributed at higher relative distances and with different overlapping ratios. The matrix $\Omega$ used in the first experiment is 
\setcounter{MaxMatrixCols}{14}
\begin{equation}
\Omega=
\begin{bmatrix}
	
	-10 & -10 &-10 &-10 &-10 &-10 &-10 &-10 &-10 &-10 & 10 & -15 & -25 & -25 \\
	-10 & 10 & -10 & 10 &-10 & 10 &-10 & 10 &-25 &-25 & 0 & 0 & -15 & 10 \\
	
	10 & -10 & 10 & -10 &10 & -10 & 20 & -10 & 15 & -15 & -10& -10& -25 & -25 \\
	
	15 & 15 &15 &15 &25&25&15 &15 &15 &15 &10&10& -15&10\\
	
	15 & 5 & -25 & 0 & -15 & 5 & -15 & 5 & -15 & 5 & 15 & 15 & 5& -10 \\
	
	-25 & 0 & 15 & 5 & 15 & 5 &15 & 5 &15 & 5 &15 & 5 & 0&0 \\
	
	5 & 15 & 5 & 15 &5 & 15 &5 & 15 &5 & 15 &10&15&-25&25 \\
	
	5 & -20 & 5 & -20 &5 & -20 &5 & -15 &5 & -15 &-15&-10&25&-25 \\
	
	-5 & -5 & -5 & -5 &-5 & -5 &-5 & -5 &-5 & -5 &15&10&25&25 \\
	
	5 & 5 & 5 & 5 &5 & 5 &5 & 5 &5 & 5 &0&0&25&0 \\
	
	-5 & 5 & -5 & 5 &-5 & 5 &-5 & 5 &-5 & 5 &-15&10&-10&10 \\
	
	5 & -5 & 5 & -5 &5 & -5 &5 & -5 &5 & -5 &25&-25&10&-10 \\
    \end{bmatrix}
.  \nonumber
\end{equation}

\begin{table}
  \centering
  \caption{Comparison of the classification performances ($\%$) of the base-layer classifiers with respect to the classes (\textit{\textbf{Class-ClassID}}) and the performances of the FSG,  when $\hat{Ave}_{corr}=1$.}
    \begin{tabular}{cccccccccccc}
    
    \toprule
      \centering
          & \textbf{$F_1$} & \textbf{$F_2$} & \textbf{$F_3$} & \textbf{$F_4$} & \textbf{$F_5$} & \textbf{$F_6$} & \textbf{$F_7$} & FSG \\
    \midrule
    \textit{\textbf{Class-1}} & {66.0$\%$} & {63.6$\%$} & {67.6$\%$} & {62.8$\%$} & {61.6$\%$} & {85.6$\%$} & {50.0$\%$} & \textbf{100.0$\%$} \\
    \textit{\textbf{Class-2}} & {67.2$\%$} & {60.8$\%$} & {49.6$\%$} & {50.8$\%$} & {\textit{\textbf{98.4$\%$}}} & {38.4$\%$} & {36.8$\%$} & \textbf{100.0$\%$} \\
    \textit{\textbf{Class-3}} & {54.4$\%$} & {58.8$\%$} & {50.8$\%$} & {\textit{\textbf{85.2$\%$}}} & {72.4$\%$} & {53.6$\%$} & {47.6$\%$} & \textbf{99.2$\%$} \\
    \textit{\textbf{Class-4}} & {66.8$\%$} & {64.0$\%$} & {\textit{\textbf{96.8$\%$}}} & {66.4$\%$} & {61.6$\%$} & {22.8$\%$} & {37.6$\%$} & \textbf{100.0$\%$} \\
    \textit{\textbf{Class-5}} & {60.8$\%$} & {\textit{\textbf{90.0$\%$}}} & {56.0$\%$} & {63.6$\%$} & {75.2$\%$} & {38.8$\%$} & {48.4$\%$} & \textbf{100.0$\%$} \\
    \textit{\textbf{Class-6}} & {\textit{\textbf{91.6$\%$}}} & {57.2$\%$} & {69.6$\%$} & {54.0$\%$} & {66.0$\%$} & {43.6$\%$} & {73.6$\%$} & \textbf{100.0$\%$} \\
    \textit{\textbf{Class-7}} & {57.2$\%$} & {55.2$\%$} & {65.2$\%$} & {57.6$\%$} & {60.8$\%$} & {37.2$\%$} & {\textit{\textbf{94.4$\%$}}} & \textbf{100.0$\%$} \\
    \textit{\textbf{Class-8}} & {78.4$\%$} & {75.6$\%$} & {86.0$\%$} & {69.2$\%$} & {54.4$\%$} & {61.6$\%$} & {\textit{\textbf{97.6$\%$}}} & \textbf{100.0$\%$} \\
    \textit{\textbf{Class-9}} & {40.8$\%$} & {41.2$\%$} & {36.0$\%$} & {36.0$\%$} & {32.8$\%$} & {26.0$\%$} & {\textit{\textbf{99.6$\%$}}} & \textbf{100.0$\%$} \\
    \textit{\textbf{Class-10}} & {44.0$\%$} & {32.4$\%$} & {32.0$\%$} & {38.0$\%$} & {37.6$\%$} & {43.2$\%$} & {\textit{\textbf{95.6$\%$}}} & \textbf{100.0$\%$} \\
    \textit{\textbf{Class-11}} & {32.0$\%$} & {35.2$\%$} & {33.6$\%$} & {40.0$\%$} & {39.6$\%$} & {\textit{\textbf{92.8$\%$}}} & {38.8$\%$} & \textbf{99.6$\%$} \\
    \textit{\textbf{Class-12}} & {37.6$\%$} & {39.6$\%$} & {34.4$\%$} & {52.0$\%$} & {44.4$\%$} & {\textit{\textbf{97.2$\%$}}} & {63.6$\%$} & \textbf{99.6$\%$} \\
    \textit{\textbf{Average Performance ($\%$)}} & {58.0$\%$} & {56.1$\%$} & {56.5$\%$} & {56.3$\%$} & {58.7$\%$} & {53.4$\%$} & {65.3$\%$} & \textbf{99.9$\%$} \\
    \bottomrule
    \end{tabular}%
  \label{tab:fsg_table1}%
\end{table}%

\newpage

In Table \ref{tab:fsg_table2}, the performance results of algorithms at another epoch of the experiments are given. In this experiment, $90\%$ of the samples are correctly classified by at least one of the base-layer classifiers, i.e. $\hat{Ave}_{corr}=0.9$. The corresponding mean value matrix $\Omega$ of each class at each feature space is 
\setcounter{MaxMatrixCols}{14}
\begin{equation}
\Omega=
\begin{bmatrix}
	
	-20 & -20 &-10 &-10 &-10 &-10 &-10 &-10 &10 &-10 & 10 & -15 & 15 & 5 \\
	-20 & 20 & -10 & 10 &-10 & 10 &-10 & 10 &-5 &-10 & 0 & 0 & -5 & 10 \\
	
	10 & -10 & 20 & -20 &10 & -10 & 10 & -10 & 15 & -15 & -10& -10& -10 & -5 \\
	
	15 & 15 &25 &25 &5&5&-5 &10 &15 &15 &10&10& -15&10\\
	
	15 & 5 & -5 & 0 & -25 & 25 & -10 & 5 & -5 & 5 & 15 & 15 & 5& -10 \\
	
	-5 & 0 & 15 & 5 & 25 & 25 &15 & 5 &15 & 5 &15 & 5 & 0&0 \\
	
	5 & 15 & 5 & 15 &5 & 15 &25 & 25 &5 & 10 &10&15&-5&5 \\
	
	5 & -20 & 5 & -10 &5 & -5 &25 & 25 &5 & -15 &-15&-10&5&-5 \\
	
	-5 & -5 & -5 & -5 &-5 & -5 &-5 & -5 &-25 & -25 &15&10&5&5 \\
	
	5 & 5 & 5 & 5 &5 & 5 &5 & 5 &25 & 25 &0&0&5&0 \\
	
	-5 & 5 & -5 & 5 &-5 & 5 &-5 & 5 &-5 & 5 &-25&25&-10&10 \\
	
	5 & -5 & 5 & -5 &5 & -5 &5 & -5 &5 & -5 &15&-15&25&-25 \\
    \end{bmatrix}
.  \nonumber
\end{equation}
\begin{table}
  \centering
  \caption{Comparison of the classification performances ($\%$) of the base-layer classifiers with respect to the classes (\textit{\textbf{Class-ClassID}}) and the performances of the FSG, when $\hat{Ave}_{corr}=0.9$.}
    \begin{tabular}{cccccccccccc}
        \toprule
          & \textbf{$F _{1}$} & \textbf{$F _{2}$} & \textbf{$F _{3}$} & \textbf{$F _{4}$} & \textbf{$F _{5}$} & \textbf{$F _{6}$} & \textbf{$F _{7}$} & FSG \\
    \midrule
    \textit{\textbf{Class-1}} & {\textit{\textbf{97.2$\%$}}} & {67.6$\%$} & {68.4$\%$} & {69.6$\%$} & {28.0$\%$} & {53.6$\%$} & {65.6$\%$} & \textbf{100.0$\%$} \\
    \textit{\textbf{Class-2}} & {\textit{\textbf{96.8$\%$}}} & {63.2$\%$} & {63.6$\%$} & {41.6$\%$} & {67.6$\%$} & {44.4$\%$} & {30.0$\%$} & \textbf{100.0$\%$} \\
    \textit{\textbf{Class-3}} & {56.4$\%$} & {\textit{\textbf{95.2$\%$}}} & {57.2$\%$} & {66.8$\%$} & {56.8$\%$} & {47.2$\%$} & {66.4$\%$} & \textbf{99.6$\%$} \\
    \textit{\textbf{Class-4}} & {60.8$\%$} & {\textit{\textbf{98.0$\%$}}} & {22.8$\%$} & {30.8$\%$} & {62.0$\%$} & {24.4$\%$} & {46.0$\%$} & \textbf{100.0$\%$} \\
    \textit{\textbf{Class-5}} & {56.8$\%$} & {24.0$\%$} & {\textit{\textbf{96.8$\%$}}} & {27.2$\%$} & {44.8$\%$} & {38.8$\%$} & {50.4$\%$} & \textbf{100.0$\%$} \\
    \textit{\textbf{Class-6}} & {32.8$\%$} & {68.4$\%$} & {\textit{\textbf{97.6$\%$}}} & {71.2$\%$} & {57.2$\%$} & {43.6$\%$} & {14.0$\%$} & \textbf{100.0$\%$} \\
    \textit{\textbf{Class-7}} & {54.0$\%$} & {65.6$\%$} & {74.4$\%$} & {\textit{\textbf{96.8$\%$}}} & {52.4$\%$} & {36.8$\%$} & {24.4$\%$} & \textbf{99.6$\%$} \\
    \textit{\textbf{Class-8}} & {77.2$\%$} & {43.6$\%$} & {29.6$\%$} & {\textit{\textbf{98.4$\%$}}} & {48.0$\%$} & {65.6$\%$} & {27.6$\%$} & \textbf{99.6$\%$} \\
    \textit{\textbf{Class-9}} & {45.2$\%$} & {34.0$\%$} & {35.2$\%$} & {35.2$\%$} & {\textit{\textbf{98.8$\%$}}} & {24.8$\%$} & {29.2$\%$} & \textbf{100.0$\%$} \\
    \textit{\textbf{Class-10}} & {40.0$\%$} & {33.6$\%$} & {22.4$\%$} & {47.6$\%$} & {\textit{\textbf{90.4$\%$}}} & {33.6$\%$} & {18.0$\%$} & \textbf{100.0$\%$} \\
    \textit{\textbf{Class-11}} & {49.2$\%$} & {28.4$\%$} & {38.0$\%$} & {28.0$\%$} & {38.4$\%$} & {\textit{\textbf{100.0$\%$}}} & {26.0$\%$} & \textbf{100.0$\%$} \\
    \textit{\textbf{Class-12}} & {34.8$\%$} & {34.4$\%$} & {22.4$\%$} & {34.4$\%$} & {44.4$\%$} & {65.2$\%$} & {\textit{\textbf{98.8$\%$}}} & \textbf{100.0$\%$} \\
    \textit{\textbf{Average Performance ($\%$)}} & {58.4$\%$} & {54.6$\%$} & {52.3$\%$} & {53.9$\%$} & {57.4$\%$} & {48.1$\%$} & {41.3$\%$} & \textbf{99.9$\%$} \\
    \bottomrule
    \end{tabular}%
  \label{tab:fsg_table2}%
\end{table}%

In the third set of the experiments, samples are distributed in the descriptors such that $80\%$ of the samples are correctly classified by at least one base-layer classifier ($\hat{Ave}_{corr}=0.8$). The performance results of the experiment are provided in Table \ref{tab:fsg_table3} and the corresponding mean value matrix $\Omega$ is 

\setcounter{MaxMatrixCols}{14}
\begin{equation}
\Omega=
\begin{bmatrix}
	
	-12 & -12 &-7.5 &-7.5 &-10 &-10 &-7.5 &-7.5 &10 &-10 & 10 & -15 & 10 & 5 \\
	-10 & 10 & -8 & 8 &-10 & 10 &-10 & 10 &-5 &-10 & 0 & 0 & -5 & 10 \\
	
	10 & -10 & 10 & -15 &10 & -10 & 10 & -10 & 10 & -15 & -10& -10& -5 & -5 \\
	
	15 & 15 &15 &17.5 &5&5&-5 &10 &15 &15 &10&10& -15&10\\
	
	15 & 5 & -5 & 0 & -15 & 15 & -10 & 5 & -5 & 5 & 15 & 15 & 5& -10 \\
	
	-5 & 0 & 15 & 5 & 15 & 15 &10 & 5 &10 & 5 &10 & 5 & 0&0 \\
	
	5 & 15 & 5 & 15 &5 & 15 &10 & 15 &5 & 10 &10&15&-5&5 \\
	
	5 & -15 & 5 & -10 &5 & -5 &15 & -15 &5 & -15 &-15&-10&5&-5 \\
	
	-5 & -5 & -5 & -5 &-5 & -5 &-5 & -5 &-10 & -15 &15&10&5&5 \\
	
	5 & 5 & 5 & 5 &5 & 5 &5 & 5 &20 & 20 &0&0&5&0 \\
	
	-5 & 5 & -5 & 5 &-5 & 5 &-5 & 5 &-5 & 5 &-5&10&-10&10 \\
	
	5 & -5 & 5 & -5 &5 & -5 &5 & -5 &5 & -5 &15&-15&10&-15 \\
    \end{bmatrix}
.  \nonumber
\end{equation}
\begin{table}
  \centering
  \caption{Comparison of the classification performances ($\%$) of the base-layer classifiers with respect to the classes (\textit{\textbf{Class-ClassID}}) and the performances of the FSG, when $\hat{Ave}_{corr}=0.8$.}
     \begin{tabular}{cccccccccccc}
     
    \toprule
          & \textbf{$F _{1}$} & \textbf{$F _{2}$} & \textbf{$F _{3}$} & \textbf{$F _{4}$} & \textbf{$F _{5}$} & \textbf{$F _{6}$} & \textbf{$F _{7}$} & FSG \\
    \midrule
    \textit{\textbf{Class-1}} & {\textit{\textbf{82.8$\%$}}} & {63.6$\%$} & {66.0$\%$} & {71.2$\%$} & {32.0$\%$} & {54.0$\%$} & {67.2$\%$} & \textbf{99.6$\%$} \\
    \textit{\textbf{Class-2}} & {\textit{\textbf{73.2$\%$}}} & {63.6$\%$} & {48.0$\%$} & {34.4$\%$} & {51.6$\%$} & {37.6$\%$} & {29.6$\%$} & \textbf{97.2$\%$} \\
    \textit{\textbf{Class-3}} & {55.2$\%$} & {\textit{\textbf{78.0$\%$}}} & {59.6$\%$} & {51.2$\%$} & {62.4$\%$} & {46.8$\%$} & {69.6$\%$} & \textbf{98.4$\%$} \\
    \textit{\textbf{Class-4}} & {61.2$\%$} & {\textit{\textbf{82.0$\%$}}} & {26.0$\%$} & {31.2$\%$} & {44.4$\%$} & {17.6$\%$} & {52.8$\%$} & \textbf{98.4$\%$} \\
    \textit{\textbf{Class-5}} & {53.2$\%$} & {23.2$\%$} & {\textit{\textbf{76.8$\%$}}} & {29.6$\%$} & {41.2$\%$} & {39.6$\%$} & {45.2$\%$} & \textbf{100.0$\%$} \\
    \textit{\textbf{Class-6}} & {24.8$\%$} & {66.4$\%$} & {\textit{\textbf{87.2$\%$}}} & {62.0$\%$} & {56.4$\%$} & {42.4$\%$} & {21.2$\%$} & \textbf{98.8$\%$} \\
    \textit{\textbf{Class-7}} & {54.0$\%$} & {63.2$\%$} & {54.8$\%$} & {\textit{\textbf{88.4$\%$}}} & {55.2$\%$} & {36.8$\%$} & {23.6$\%$} & \textbf{98.4$\%$} \\
    \textit{\textbf{Class-8}} & {\textit{\textbf{80.8$\%$}}} & {39.2$\%$} & {22.8$\%$} & {74.8$\%$} & {45.2$\%$} & {63.2$\%$} & {23.6$\%$} & \textbf{96.4$\%$} \\
    \textit{\textbf{Class-9}} & {39.6$\%$} & {33.2$\%$} & {33.2$\%$} & {29.6$\%$} & {\textit{\textbf{83.6$\%$}}} & {21.6$\%$} & {29.6$\%$} & \textbf{99.2$\%$} \\
    \textit{\textbf{Class-10}} & {38.4$\%$} & {35.6$\%$} & {30.8$\%$} & {47.6$\%$} & {\textit{\textbf{82.8$\%$}}} & {38.0$\%$} & {24.0$\%$} & \textbf{99.2$\%$} \\
    \textit{\textbf{Class-11}} & {33.2$\%$} & {30.0$\%$} & {30.8$\%$} & {30.4$\%$} & {38.8$\%$} & {\textit{\textbf{84.4$\%$}}} & {29.6$\%$} & \textbf{96.4$\%$} \\
    \textit{\textbf{Class-12}} & {40.4$\%$} & {33.2$\%$} & {28.0$\%$} & {40.4$\%$} & {32.4$\%$} & {58.8$\%$} & {\textit{\textbf{81.2$\%$}}} & \textbf{99.2$\%$}\\
    \textit{\textbf{Average Performance ($\%$)}} & {53.1$\%$} & {50.9$\%$} & {47.0$\%$} & {49.2$\%$} & {52.2$\%$} & {45.1$\%$} & {41.4$\%$} & \textbf{98.4$\%$} \\
    \bottomrule
    \end{tabular}%
  \label{tab:fsg_table3}%
\end{table}%

In the fourth set of the experiments given in Table \ref{tab:fsg_table4}, samples are distributed in the descriptors such that each classifier can correctly classify $70\%$ of the samples. The corresponding mean value matrix $\Omega$ is 
\setcounter{MaxMatrixCols}{14}
\begin{equation}
\Omega=
\begin{bmatrix}
	
	-12.5 & -12.5 &-10 &-10 &-10 &-10 &-10 &-10 &10 &-10 & 10 & -15 & 15 & 5 \\
	-10 & 15 & -10 & 10 &-10 & 10 &-10 & 10 &-5 &-10 & 0 & 0 & -5 & 10 \\
	
	10 & -10 & 15 & -15 &10 & -10 & 10 & -10 & 15 & -15 & -10& -10& 10 & -5 \\
	
	15 & 15 &19 &19 &5&5&-5 &10 &15 &15 &10&10& -15&10\\
	
	15 & 5 & -5 & 0 & -17.5 & 17.5 & -10 & 5 & -5 & 5 & 15 & 15 & 5& -10 \\
	
	-5 & 0 & 15 & 5 & 17.5 & 17.5 &15 & 5 &15 & 5 &15 & 5 & 0&0 \\
	
	5 & 15 & 5 & 15 &5 & 15 &17.5 & 17.5 &5 & 10 &10&15&-5&5 \\
	
	5 & -20 & 5 & -10 &5 & -5 &17.5 & -17.5 &5 & -15 &-15&-10&5&-5 \\
	
	-5 & -5 & -5 & -5 &-5 & -5 &-5 & -5 &-15 & -15 &15&10&5&5 \\
	
	5 & 5 & 5 & 5 &5 & 5 &5 & 5 &22.5 & 22.5 &0&0&5&0 \\
	
	-5 & 5 & -5 & 5 &-5 & 5 &-5 & 5 &-5 & 5 &-10&10&-10&10 \\
	
	5 & -5 & 5 & -5 &5 & -5 &5 & -5 &5 & -5 &15&-15&15&-15 \\
    \end{bmatrix}
. \nonumber
\end{equation}

\begin{table}
  \centering
  \caption{Comparison of the classification performances ($\%$) of the base-layer classifiers with respect to the classes (\textit{\textbf{Class-ClassID}}) and the performances of the FSG, when $\hat{Ave}_{corr}=0.7$.}
    \begin{tabular}{cccccccccccc}
    
    \toprule
          & \textbf{$F _{1}$} & \textbf{$F _{2}$} & \textbf{$F _{3}$} & \textbf{$F _{4}$} & \textbf{$F _{5}$} & \textbf{$F _{6}$} & \textbf{$F _{7}$} & FSG \\
    \midrule
    \textit{\textbf{Class-1}} & {\textit{\textbf{75$\%$}}} & {42$\%$} & {68$\%$} & {52$\%$} & {36$\%$} & {62$\%$} & {46$\%$} &  \textbf{99$\%$} \\
    \textit{\textbf{Class-2}} & {\textit{\textbf{64$\%$}}} & {45$\%$} & {41$\%$} & {38$\%$} & {43$\%$} & {37$\%$} & {32$\%$} & \textbf{98$\%$} \\
    \textit{\textbf{Class-3}} & {46$\%$} & {\textit{\textbf{72$\%$}}} & {60$\%$} & {40$\%$} & {39$\%$} & {52$\%$} & {46$\%$} & \textbf{88$\%$} \\
    \textit{\textbf{Class-4}} & {68$\%$} & {\textit{\textbf{72$\%$}}} & {23$\%$} & {33$\%$} & {45$\%$} & {17$\%$} & {59$\%$} & \textbf{98$\%$} \\
    \textit{\textbf{Class-5}} & {54$\%$} & {22$\%$} & {\textit{\textbf{70$\%$}}} & {28$\%$} & {40$\%$} & {42$\%$} & {32$\%$} & \textbf{100$\%$} \\
    \textit{\textbf{Class-6}} & {22$\%$} & {68$\%$} & {\textit{\textbf{74$\%$}}} & {50$\%$} & {46$\%$} & {28$\%$} & {18$\%$} & \textbf{97$\%$} \\
    \textit{\textbf{Class-7}} & {65$\%$} & {62$\%$} & {50$\%$} & {\textit{\textbf{72$\%$}}} & {44$\%$} & {34$\%$} & {20$\%$} & \textbf{96$\%$} \\
    \textit{\textbf{Class-8}} & {55$\%$} & {30$\%$} & {25$\%$} & {\textit{\textbf{75$\%$}}} & {44$\%$} & {61$\%$} & {18$\%$} & \textbf{89$\%$} \\
    \textit{\textbf{Class-9}} & {36$\%$} & {24$\%$} & {36$\%$} & {30$\%$} & {\textit{\textbf{67$\%$}}} & {32$\%$} & {23$\%$} & \textbf{100$\%$} \\
    \textit{\textbf{Class-10}} & {42$\%$} & {32$\%$} & {24$\%$} & {27$\%$} & {\textit{\textbf{74$\%$}}} & {32$\%$} & {21$\%$} & \textbf{98$\%$} \\
    \textit{\textbf{Class-11}} & {31$\%$} & {17$\%$} & {34$\%$} & {16$\%$} & {38$\%$} & {\textit{\textbf{70$\%$}}} & {26$\%$} & \textbf{95$\%$} \\
    \textit{\textbf{Class-12}} & {33$\%$} & {28$\%$} & {27$\%$} & {41$\%$} & {38$\%$} & {67$\%$} & {\textit{\textbf{68$\%$}}} & \textbf{100$\%$} \\
    \textit{\textbf{Average Performance ($\%$)}} & {49.3$\%$} & {42.9$\%$} & {44.3$\%$} & {41.8$\%$} & {46.1$\%$} & {44.4$\%$} & {34.2$\%$} & \textbf{96.4$\%$} \\
    \bottomrule
    \end{tabular}%
  \label{tab:fsg_table4}%
\end{table}%

Note that, the performance of the overall FSG decreases as the percentage of the samples that are correctly classified by at least one classifier decreases, i.e. $\hat{Ave}_{corr}$ decreases. This observation is due to the results given in the previous sections which state that the large-sample classification error of the meta-layer classifier of the FSG is bounded by Bayes Error, which can be achieved if each sample is correctly classified by at least one base-layer classifier such that the features of samples belonging to the same class reside in the same Voronoi regions in the fusion space. This observation is analyzed in the next subsection.

\subsubsection{Geometric Analysis of Feature, Decision and Fusion Spaces on Artificial Datasets}

In the FSG, membership values of the samples lie on the surface of a simplex in the $C$-dimensional decision space of each base-layer classifier. In practice, the entry of the vector $\bar{\mu }(\bar{x}_{j} )$ with the highest membership value shows the estimated class label $\hat{y}_{j}$ of a sample $s$ in $F_j$, $\forall j=1,\ldots,J$, and the membership vector of a correctly classified sample is expected to accumulate around the \textit{correct} vertex of the simplex. Concatenation operation creates a $CJ$-dimensional fusion space at the input of the meta-layer classifier in which the membership values lie on the $CJ$-dimensional simplex. The membership values of the correctly classified samples, this time, form even a more compact cluster around each vertex of the simplex, and misclassified samples are scattered all over the surface. 

\begin{figure}[ht!]
\centering
\subfloat[$F_1$]{\includegraphics*[width=3.6in,height=3.0in]{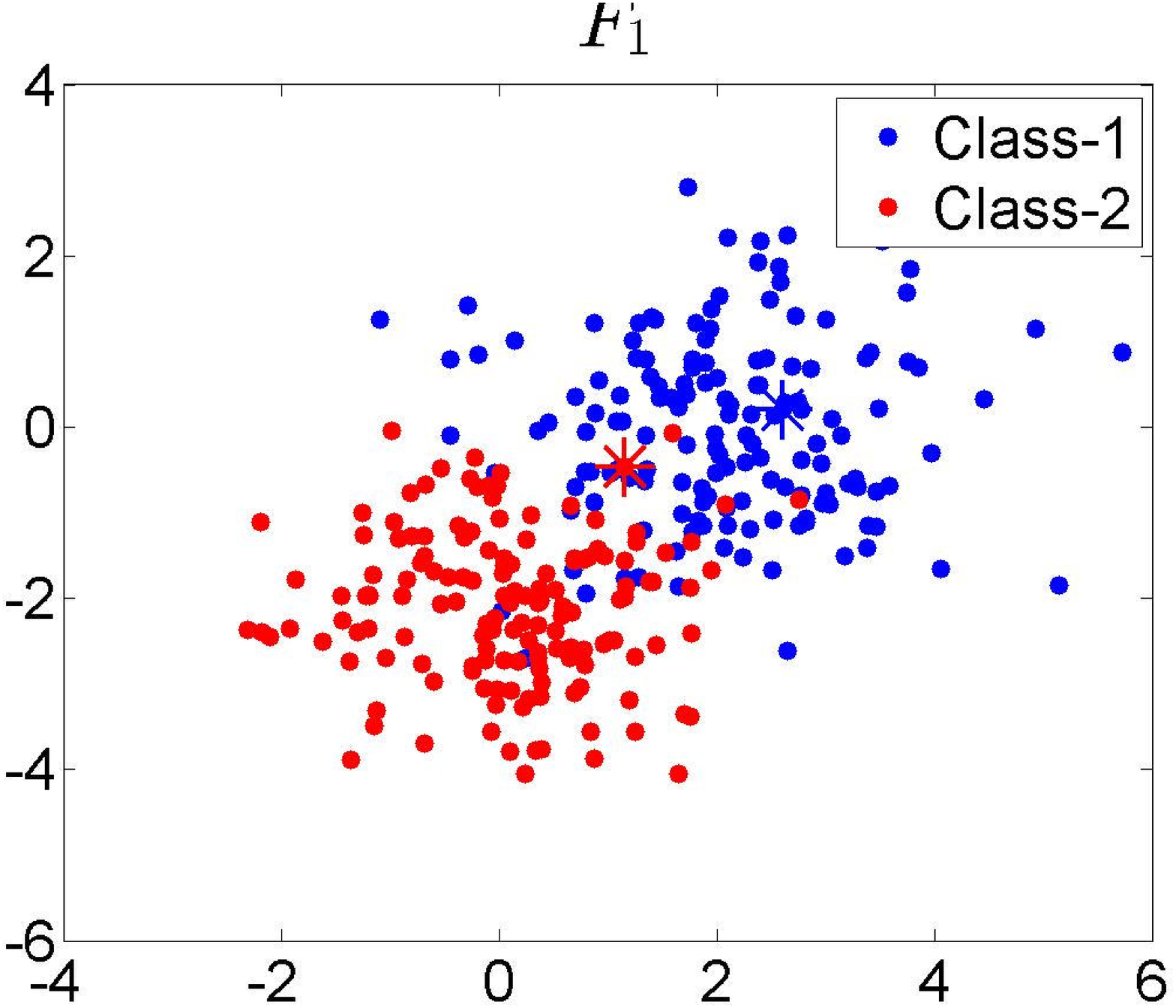}}
\subfloat[$F_2$]{\includegraphics*[width=3.6in,height=3.0in]{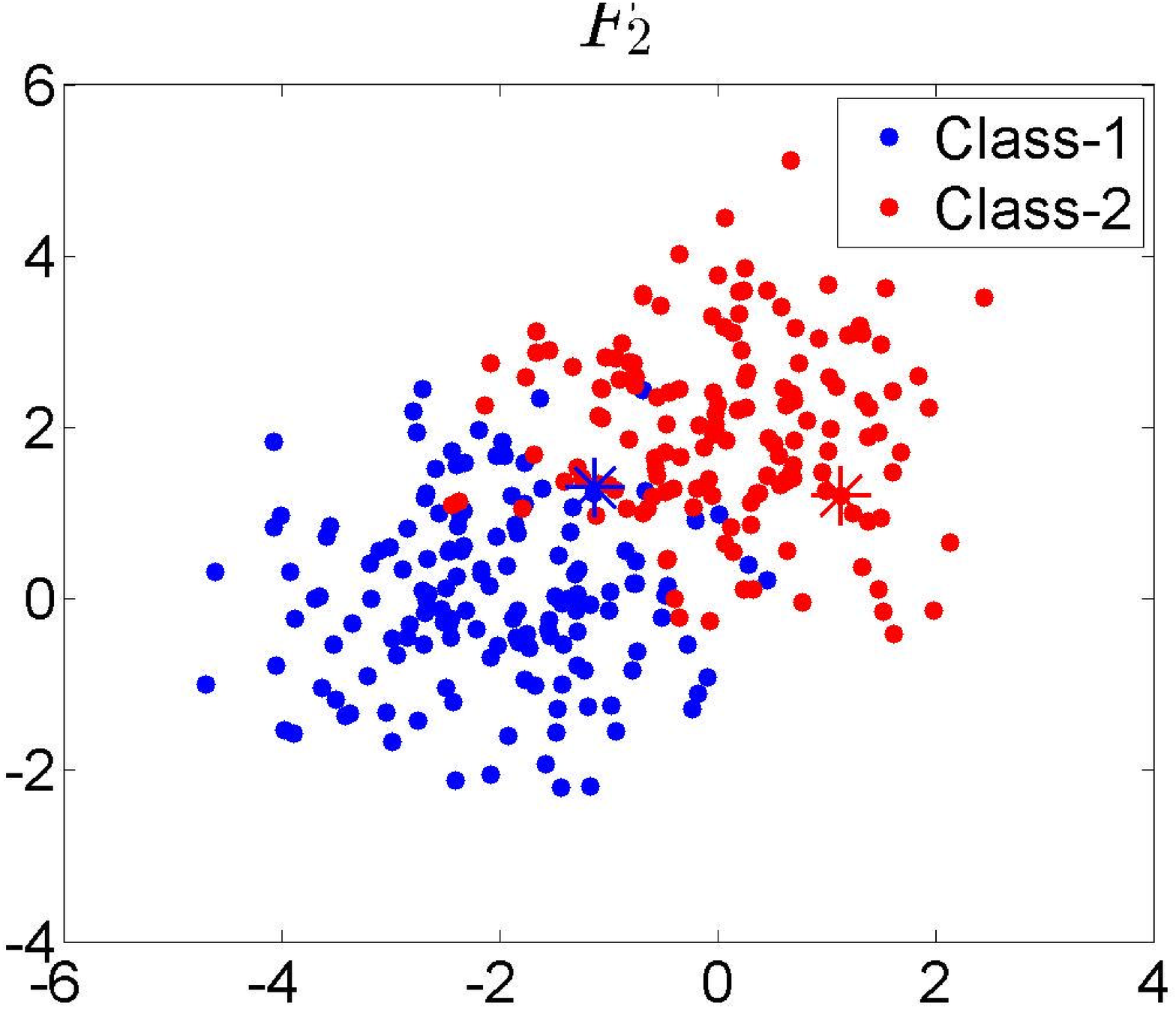}}
\caption{Feature vectors in (a) $F_1$ and (b) $F_2$. Features of two randomly selected samples are indicated by ($\ast$) to follow them at the decision spaces of base-layer classifiers and the fusion space of meta-layer classifier.}
\label{fig:fig2}
\end{figure}

Consider an artificial dataset, consisting of $C=2$ classes each of which consists of $250$ samples represented in $J=2$ distinct feature spaces. In the base-layer feature spaces shown in Fig. \ref{fig:fig2}, the classes have Gaussian distribution with substantial overlaps where the mean value and covariance matrices are
\[\Omega_{1} =\left(\begin{array}{cc} {2} & {0} \\ {0} & {-2} \end{array}\right),{\rm \; } \Sigma_{1} =\left(\begin{array}{cc} {1} & {1} \\ {1} & {1} \end{array}\right){\rm \; and\; } \Omega_{2} =\left(\begin{array}{cc} {-2} & {0} \\ {2} & {2} \end{array}\right),{\rm \; } \Sigma_{2} =\left(\begin{array}{cc} {1} & {1} \\ {1} & {1} \end{array}\right){\rm \; }\] 
for the first and the second feature spaces, respectively. The features of the samples from the first class are represented by blue dots and that of the second class are represented by red dots. Features of two randomly selected samples, which are misclassified by one of the base-layer classifiers and correctly classified by the meta-layer classifier, are shown by star ($\ast$) markers. In the feature spaces, each of the training samples is correctly classified by at least one base-layer fuzzy $k$-NN classifier with $k=3$. The classification performances of the base-layer classifiers are $91\%$ and $92\%$ respectively. The classification performance of the FSG is $96\%$. 

\begin{figure}[ht!] 
\centering
\subfloat[ ]{\includegraphics[width=3.6in,height=3.0in]{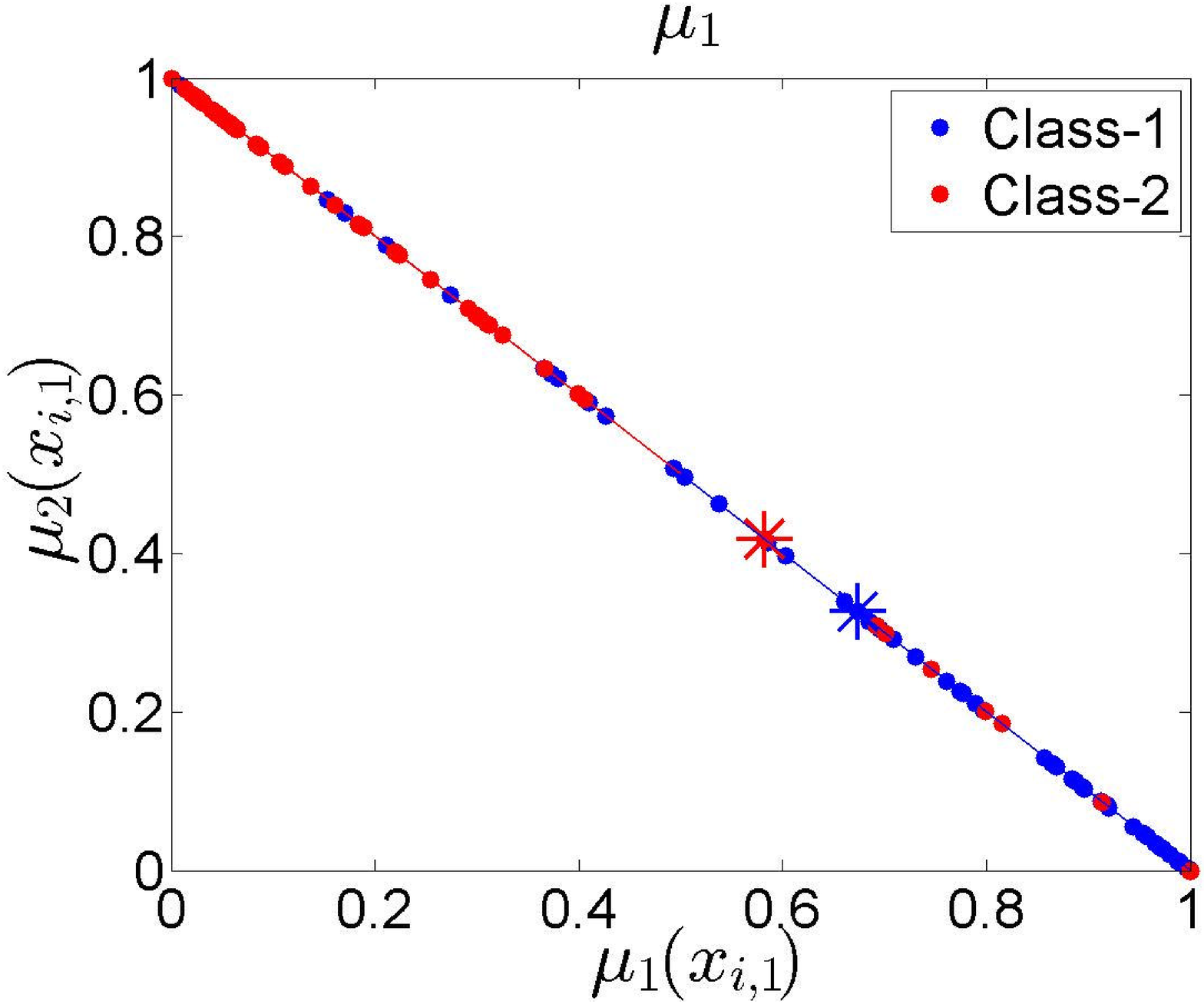}}
\subfloat[ ]{\includegraphics[width=3.6in,height=3.0in]{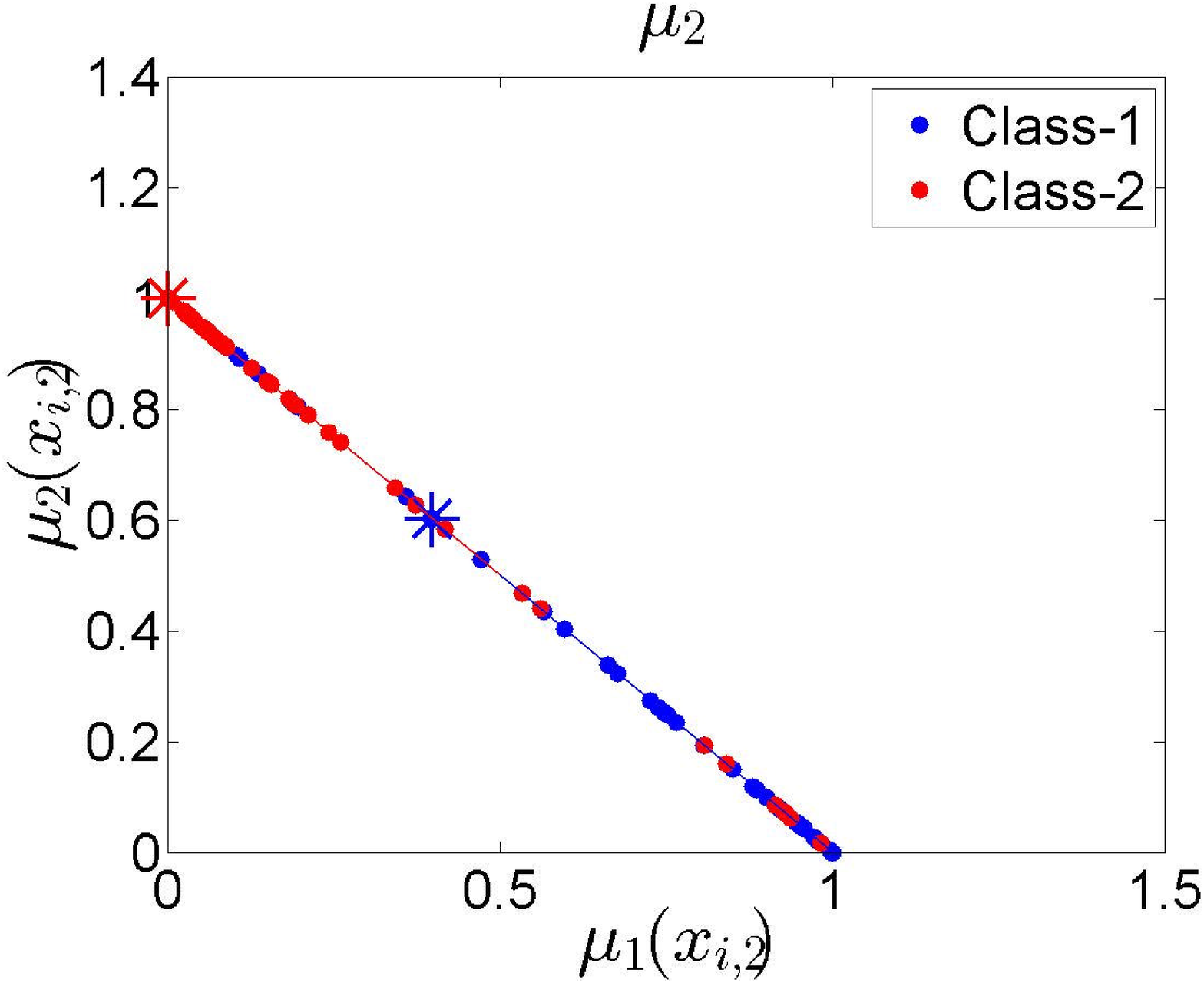}}
\caption{Membership vectors obtained at the decision spaces of base-layer classifiers: (a) Classifier 1 and (b) Classifier 2. The locations of the features of randomly selected samples of Fig.~\ref{fig:fig2} are indicated by ($\ast$), at each simplex.}
\label{fig:fig3}
\end{figure}

\begin{figure}[ht!]
\centering
\subfloat[]{\includegraphics[width=3.5in,height=2.5in]{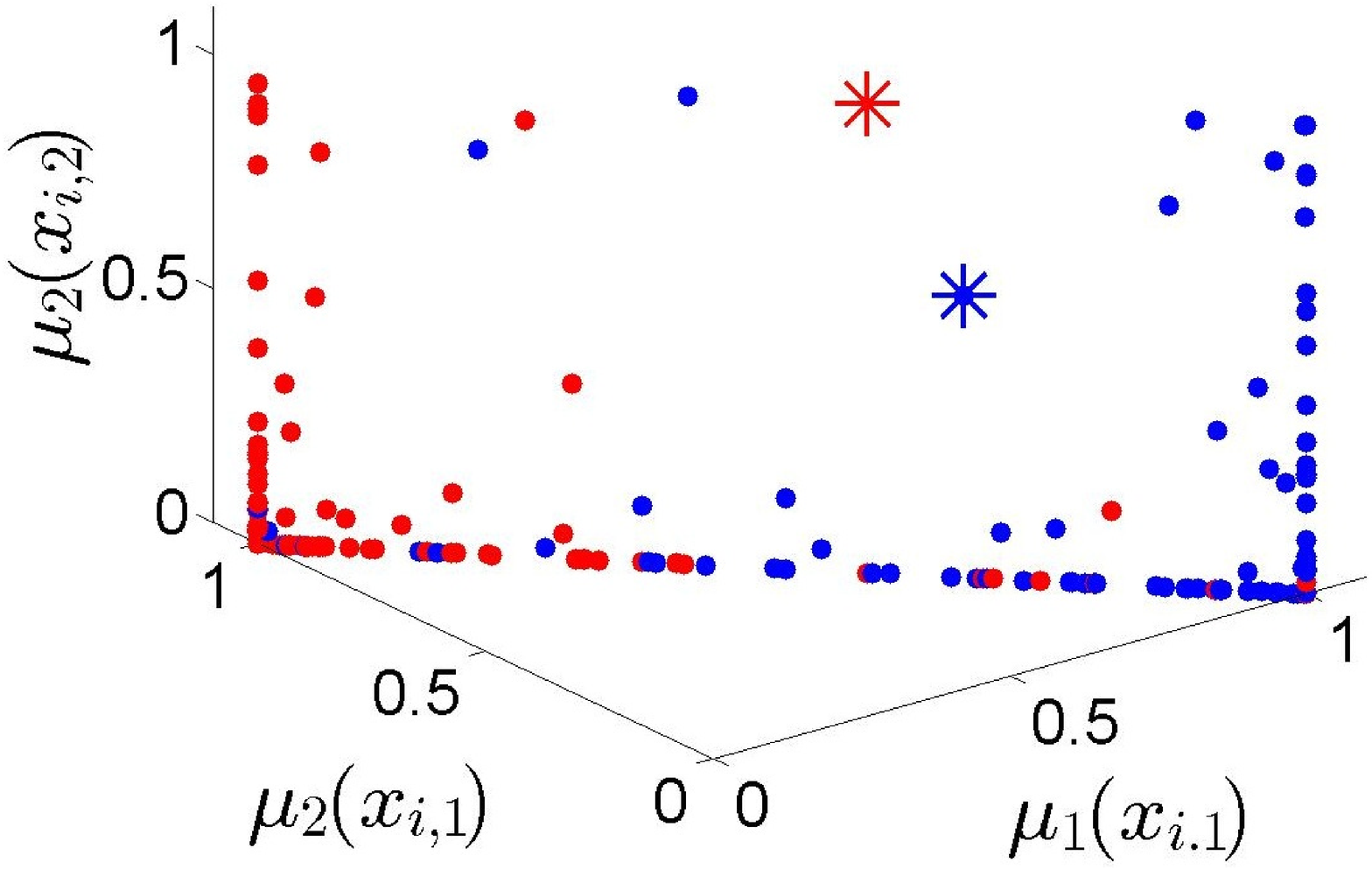}}
\subfloat[]{\includegraphics[width=3.5in,height=2.5in]{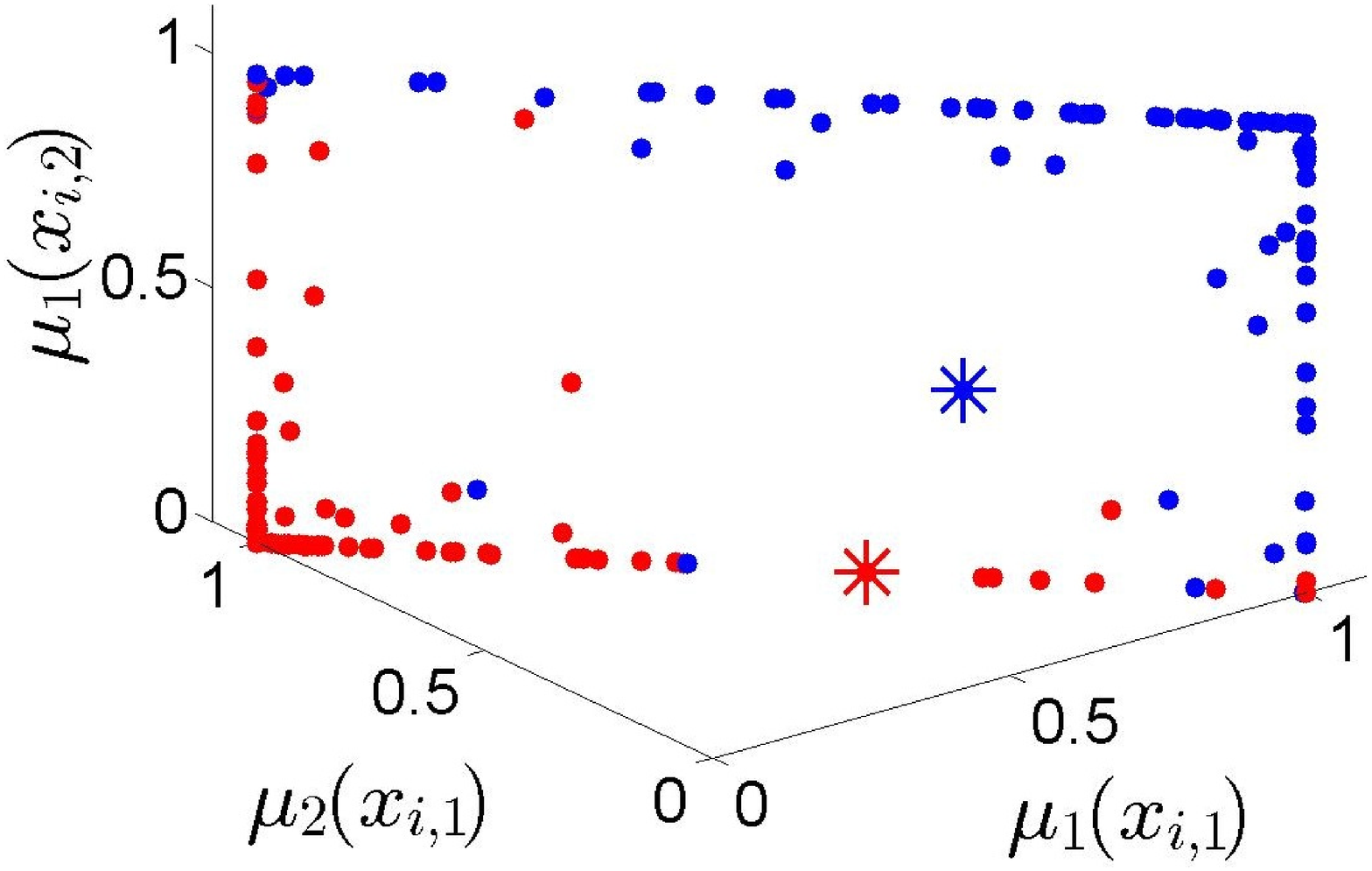}} \\
\subfloat[]{\includegraphics[width=3.5in,height=2.5in]{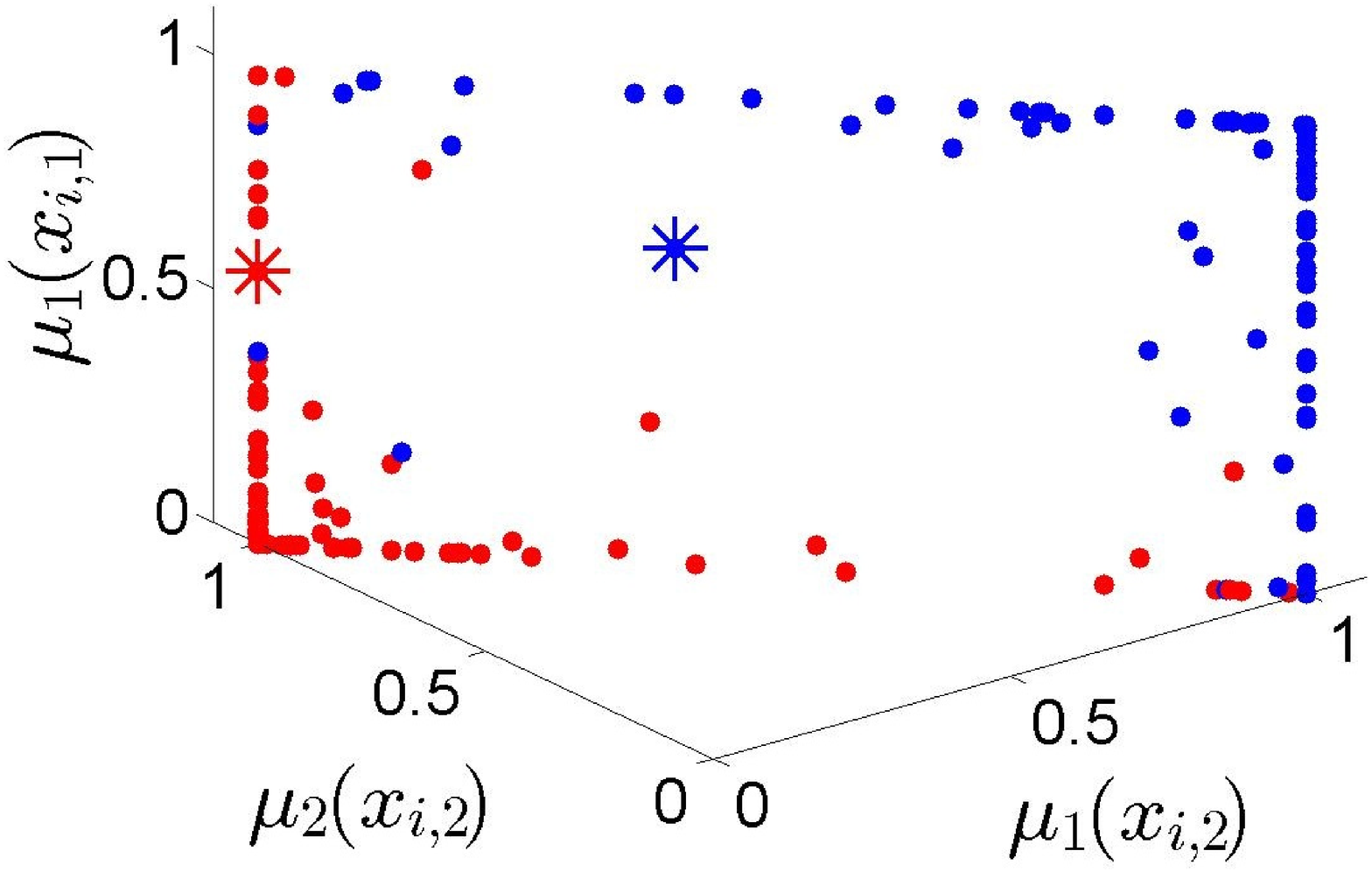}}
\subfloat[]{\includegraphics[width=3.5in,height=2.5in]{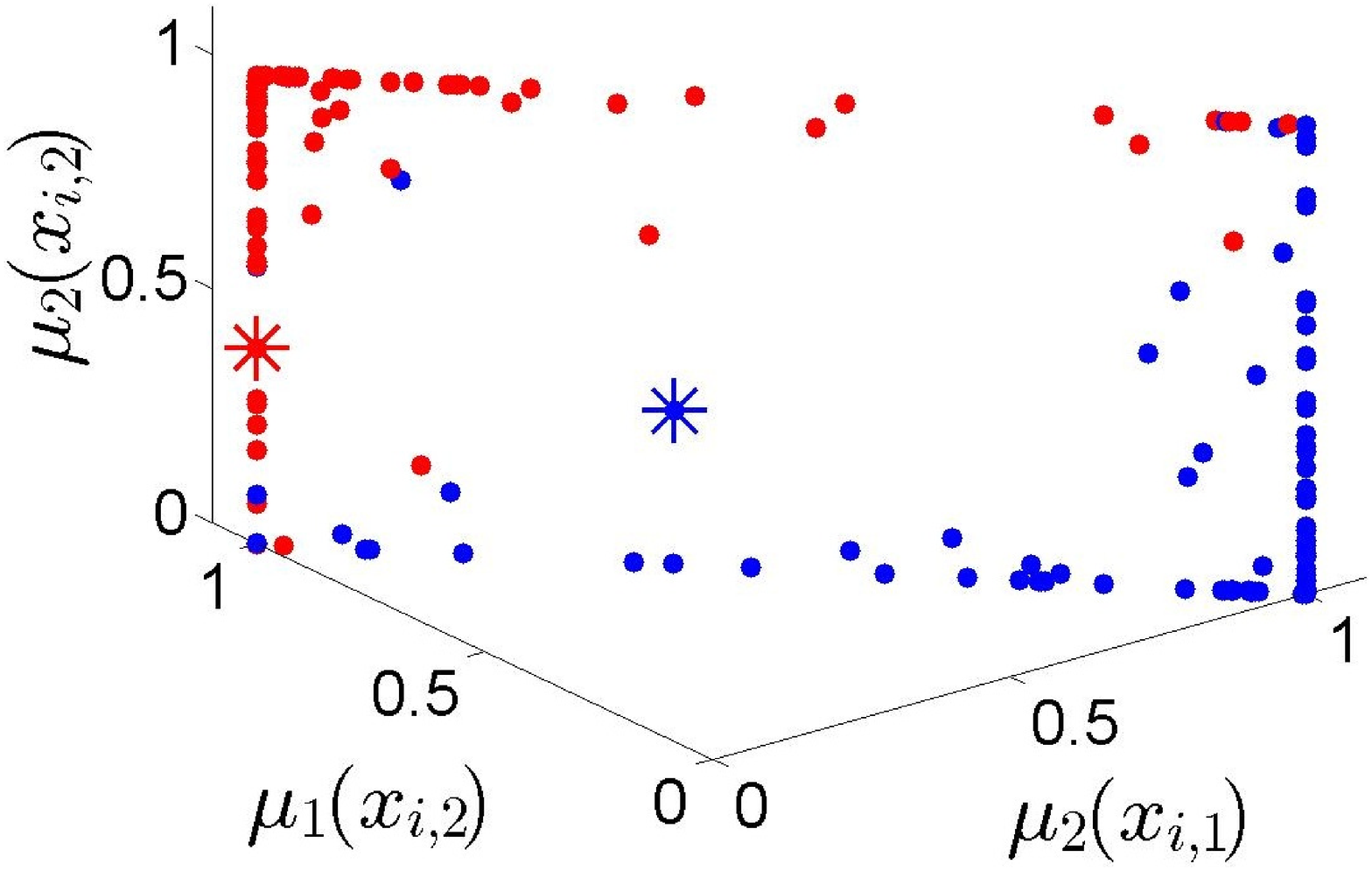}} 
\caption{The relationships among (a) $\mu_1(x_i,1)$, $ \mu_2(x_i,1)$, $\mu_2(x_i,2)$, (b) $\mu_1(x_i,1)$, $\mu_2(x_i,1)$, $\mu_1(x_i,2)$, (c) $\mu_1(x_i,2), \mu_(x_i,2),\mu_1(x_i,1)$, and (d) $\mu_2(x_i,1)$, $\mu_1(x_,2)$, $\mu_2(x_i,2)$, are visualized. The locations of the features of randomly selected samples of Fig.~\ref{fig:fig2} are indicated by ($\ast$) in the subspaces of the fusion space. }
\label{fig:fig4}
\end{figure}

The membership values lie on a line in the decision spaces of two base-layer classifiers, as depicted in Fig.~\ref{fig:fig3}. In these figures, the decisions of the classifiers are also depicted for individual samples. For instance, the sample marked with red star $s_1$ is misclassified by the first classifier as shown in Fig.~\ref{fig:fig3}.a, but correctly classified by the second classifier as shown in Fig.~\ref{fig:fig3}.b. In addition, the feature of the sample marked with blue star $s_2$ is correctly classified by the first classifier as shown in Fig.~\ref{fig:fig3}.a, but  misclassified by the second classifier as shown in Fig.~\ref{fig:fig3}.b.

The concatenation operation creates a $4$ ($2 \times 2$) dimensional fusion space at the meta-layer input feature space, i.e. fusion space. In order to visualize the distribution of concatenated membership vectors of samples in the fusion space, four different subspaces, each of which is a $3$-dimensional Euclidean space, are selected. Fig.~\ref{fig:fig4} displays different combinations of the subspaces and the membership vectors obtained from each classifier. Notice that the concatenation operation forms planes in these subspaces accumulating the correctly classified samples around the edges and the vertices. Therefore, features of the samples which are correctly classified by at least one base-layer classifier are located close to one of the \textit{correct} vertices, or edges. This fact is depicted in Fig.~\ref{fig:fig4}, where the feature of the sample indicated by red star is located close to the edges of the second class in Fig.~\ref{fig:fig4}.b, c, d. On the other hand, the feature of the sample indicated by blue star is located close to the edges of the first class in Fig.~\ref{fig:fig4}.a, c, d. Both of these samples are correctly labeled by the meta-layer fuzzy $k$-NN classifier.

\subsection{Experiments on Benchmark Datasets}
\label{sec:bench_fsg}
In the experiments, classification performances of $k=1$ nearest neighbor rule, Fuzzy Stacked Generalization (FSG), and the state of the art algorithms, Adaboost, Random Subspace (RS) and Rotation Forest (RF), are compared using benchmark datasets.

Experiments on the benchmark datasets are performed in two groups:
\begin{enumerate}
\item \textbf{Multi-attribute Datasets}: Feature vectors consisting of multiple attributes reside in a single feature space $F_j=F_{1j} \times \ldots \times F_{aj} \times \ldots \times F_{Aj}$, where $A$ is the number of attributes. In these experiments, FSG is implemented by employing individual base-layer classifiers on a feature space $F_{aj}$ consisting of an each individual attribute. Therefore, the dimension of the feature vectors in the fusion space of FSG is $CA$.

\item \textbf{Multi-feature Datasets}: Each base-layer classifier of FSG is employed on an individual feature space $F_j$, $\forall j=1,2,\ldots,J$. Therefore, the dimension of the feature vectors in the fusion space of FSG is $CJ$.

\end{enumerate}

State of the art algorithms are employed on an aggregated feature space  $F=F_{1} \times \ldots \times F_{j} \times \ldots \times F_{J}$ which contains feature vectors with dimension $A$ and $D=\sum _{j=1} ^J D_j$ in multi-attribute and multi-feature experiments, respectively.

\subsubsection{Experiments on Multi-attribute Datasets}

In the experiments, two-class Breast Cancer (BCancer), Diabetis, Flare Solar (FSolar), Thyroid, German, Titanic \cite{march,li,garc,w,blake} datasets are used as multi-attribute datasets. The number of attributes of the feature vectors of the samples in the datasets are given in Table \ref{tab:fsg_bench_att}. Training and test datasets are randomly selected from the datasets using the data splitting scheme of \cite{march,march2}. The experiments are repeated $100$ times and the average performance values are given in Table \ref{tab:fsg_bench_att}.

An interesting observation from Table \ref{tab:fsg_bench_att} is that the  $k=1$ nearest neighbor rule outperforms various well-known ensemble learning algorithms such as Adaboost and Rotation Forest, if the number of attributes is small, e.g. $A=3$. In other words, classification performance of the nearest neighbor rule decreases as $A$ increases. This observation is due to the curse of dimensionality problem of the nearest neighbor algorithms \cite{duda}. Since the dimension of the feature vectors in the fusion space is $CA$, the dimensionality curse can be observed in the fusion space of the FSG as $A$ increases. Therefore, we observe that the performance gain of FSG compared to the classification performances of the aforementioned algorithms decreases as $A$ increases. Since the dimension of the feature space of state of the art algorithms is $A$, the dimension of the fusion space of FSG is greater that of the feature space by a factor of $C$. However, the decrease of the difference between performances of FSG and the state of the art algorithms is not constant for different datasets with different $A$. Therefore, we further analyze the relationship between classification performances and the number of classes and classifiers in the next subsection.

\begin{table}[htbp]
  \centering
  \caption{Classification performances of the algorithms on Multi-attribute Datasets.}
  \begin{center}
    \begin{tabular}{ccccccc}
    \toprule
    \textbf{Datasets} & \textbf{Titanic} & \textbf{Thyroid} & \textbf{Diabetis} & \textbf{FSolar} & \textbf{BCancer} & \textbf{German}  \\
    \midrule
    \textbf{Num. of Att.}($A$) & 3 & 5  & 8 & 9  & 9   & 20  \\
    \textbf{Adaboost} & 75.06\% & 93.10\% & 75.98\% & 66.21\% & 74.87\% & 75.89\%     \\
    \textbf{Rotation Forest} & 70.14\% & 95.64\% & 72.43\% & 62.75\% & 70.58\%  & 74.81\%    \\
    \textbf{Random Subspace} & 74.83\% & 94.78\% & 74.40\% & 65.04\% & 74.08\% & 75.17\%    \\
    \textbf{1 NN} & 75.54\% & 95.64\% & 69.88\% & 60.58\% & 67.30\% & 71.12\%   \\
    \textbf{FSG} & 76.01\% & 96.41\% & 77.42\% & 67.33\% & 75.51\% & 75.30\%   \\   
    \bottomrule
    \end{tabular}%
  \end{center}
  \label{tab:fsg_bench_att}%
\end{table}%

\subsubsection{Experiments on Multi-feature Datasets}

In this section, the algorithms have been analyzed on two image classification benchmark datasets, which are Corel Dataset\footnote{The dataset is available on https://github.com/meteozay/Corel\_Dataset.git} consisting of $599$ classes and Caltech 101 Dataset consisting of 102 classes.

\textbf{7.2.2.1 Experiments on Corel Dataset}

In the Corel Dataset experiments, $4$ to $8$ feature (descriptor) combinations of Haar and 7 of MPEG-7 visual features (descriptors) \cite{mpeg1,mpeg2} features are used over $10$ to $30$ classes, each of which contains $97-100$ samples from the dataset. $50$ of the samples of each class are used for the training, and the remaining samples are used for testing. 

The feature set combinations are selected as following: 
\begin{itemize}

\item	\textbf{4 Features (4FS)}: Color Structure, Color Layout, Edge Histogram, Region-based Shape, 
\item	\textbf{5 Features (5FS)}: Color Structure, Color Layout, Edge Histogram, Region-based Shape, Haar, 
\item	\textbf{6 Features (6FS)}: Color Structure, Color Layout, Edge Histogram, Region-based Shape, Haar, Dominant Color, 
\item	\textbf{7 Features (7FS)}: Color Structure, Color Layout, Edge Histogram, Region-based Shape, Haar, Dominant Color, Scalable Color,  and 
\item	\textbf{8 Features (8FS)}: Color Structure, Color Layout, Edge Histogram, Region-based Shape, Haar, Dominant Color, Scalable Color, Homogenous Texture. 
\end{itemize}

The selected MPEG-7 features have high variance and a well-balanced cluster structure \cite{mpeg1}. These properties allow us to distinguish the samples in different classes. In addition, the feature vectors in the descriptors satisfy i.i.d. (independent and identically distribution) conditions by providing high between class variance values \cite{mpeg1}. Therefore, the statistical properties of the feature spaces provide wealthy information variability. 

In the Corel Dataset, two types of experiments are employed. In the first type of the experiments, samples belonging to a set of pre-defined classes is selected to construct smaller datasets. Then the change of the performance of the algorithms is analyzed as new samples belonging to new classes are added to the datasets and new features are added to the feature sets. In the second type of the experiments, the datasets are constructed by selecting samples belonging to randomly selected classes. In these experiments, the random class selection procedure is repeated $10$ times and the average performance is given.

The pre-defined class names of 10, 15 and 20 class classification experiments are the following

\begin{itemize}
\item \textbf{10 Class Classification}: New Guinea, Beach, Rome, Bus, Dinosaurs, Elephant, Roses, Horses, Mountain, and Dining,
\item \textbf{15 Class Classification}: Classes used in \textbf{10 Class Classification} together with Autumn, Bhutan, California Sea, Canada Sea and Canada West,
\item \textbf{20 Class Classification}: Classes used in \textbf{15 Class Classification} together with China, Croatia, Death Valley, Dogs and England.
\end{itemize}

When the sample set is fixed, the change of the classification performance is analyzed as the new features are added from combinations of \textbf{4FS} to \textbf{8FS} feature sets. The classification results given in Table \ref{tab:corel_various} show that FSG outperforms the benchmark algorithms. Moreover, the performances of the algorithms which employ majority voting to the classifier decision may decrease as new features are added. For instance, when Dominant Color and Scalable Color features are added to the combination of features in \textbf{5FS} to construct \textbf{6FS} and \textbf{7FS}, the classification performances of the FSG and the Random Subspace, which employ majority voting at the meta-layer classifiers, decrease. 



\begin{table}[htbp]
  \centering
  \caption{Classification results on the Corel Dataset with varying number of features and classes.}
    \begin{tabular}{rcccccc}
    \toprule
          & \textbf{Algorithms} & \textbf{4FS} & \textbf{5FS} & \textbf{6FS} & \textbf{7FS} & \textbf{8FS} \\
    \midrule
    {\multirow{5}{*}{10-Class Experiments}} & \textbf{Adaboost} & 63.0\% & 63.6\% & 63.2\% & 66.6\% & 67.2\% \\
    {} & \textbf{Rotation Forest} & 76.2\% & 74.4\% & 74.6\% & 76.6\% & 78.2\% \\
    {} & \textbf{Random Subspace} & 78.1\% & 77.5\% & 75.8\% & 76.9\% & 75.5\% \\
    {} & \textbf{FSG} & 85.6\% & 86.8\% & 85.6\% & 85.8\% & 85.8\% \\
          & \multicolumn{1}{r}{} & \multicolumn{1}{r}{} & \multicolumn{1}{r}{} & \multicolumn{1}{r}{} & \multicolumn{1}{r}{} & \multicolumn{1}{r}{} \\
    {\multirow{5}{*}{15-Class Experiments}} & \textbf{Adaboost} & 42.2\% & 45.5\% & 43.2\% & 46.8\% & 46.8\% \\
    {} & \textbf{Rotation Forest} & 60.2\% & 60.6\% & 60.9\% & 60.9\% & 61.3\% \\
    {} & \textbf{Random Subspace} & 65.5\% & 64.1\% & 59.8\% & 63.3\% & 61.8\% \\
    {} & \textbf{FSG} & 66.2\% & 65.3\% & 62.3\% & 62.8\% & 64.5\% \\
          & \multicolumn{1}{r}{} & \multicolumn{1}{r}{} & \multicolumn{1}{r}{} & \multicolumn{1}{r}{} & \multicolumn{1}{r}{} & \multicolumn{1}{r}{} \\
    {\multirow{5}{*}{20-Class Experiments}} & \textbf{Adaboost} & 23.3\% & 27.0\% & 27.0\% & 27.0\% & 27.0\% \\
    {} & \textbf{Rotation Forest} & 47.7\% & 49.5\% & 49.5\% & 49.6\% & 50.4\% \\
    {} & \textbf{Random Subspace} & 48.3\% & 48.1\% & 48.1\% & 48.6\% & 48.7\% \\
    {} & \textbf{FSG} & 52.4\% & 50.7\% & 49.9\% & 50.9\% & 52.9\% \\
    \bottomrule
    \end{tabular}%
  \label{tab:corel_various}%
\end{table}%

\begin{figure}[htbp]

\includegraphics[width=7.0in,height=8.8in]{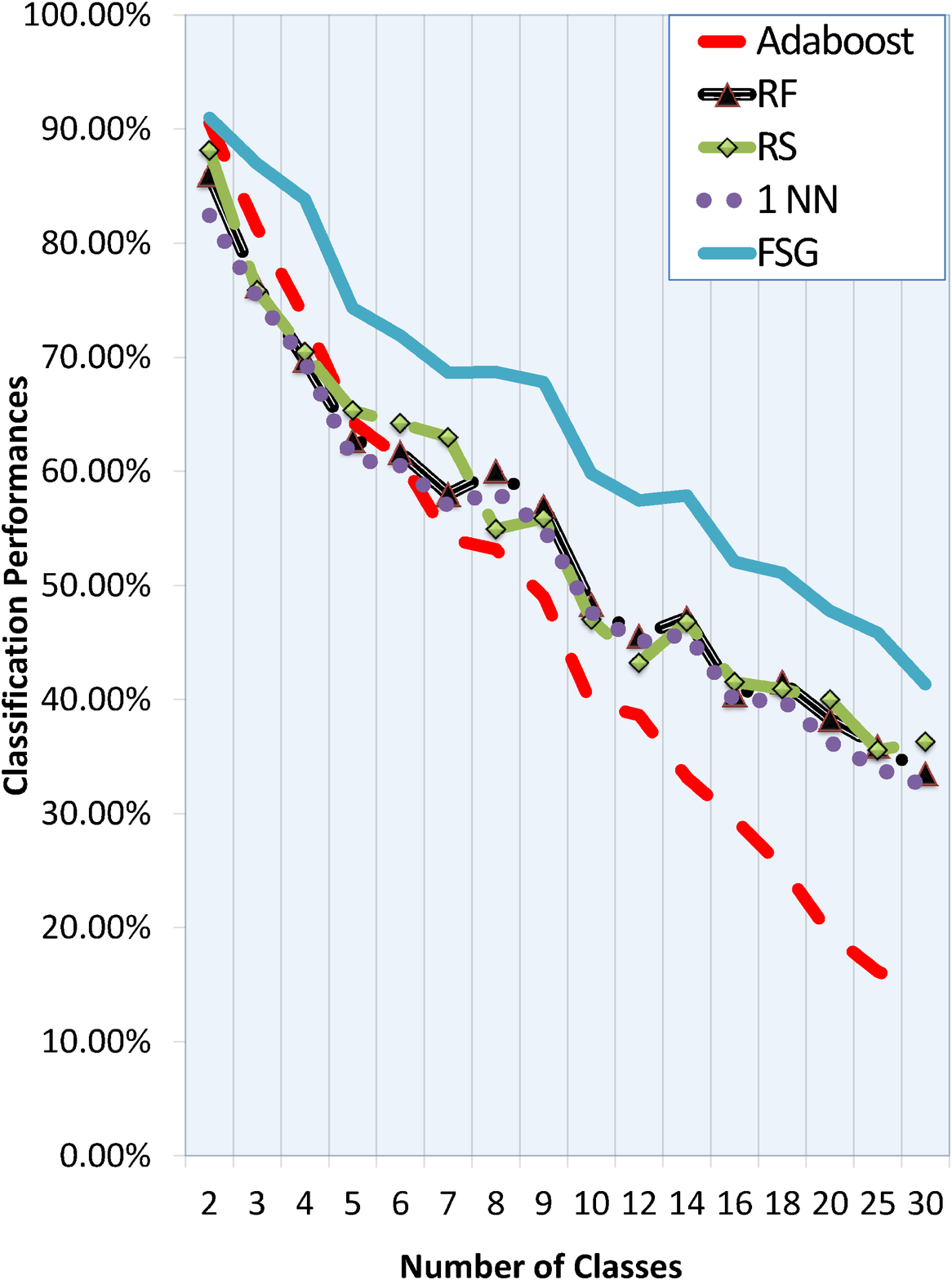}
\caption{Classification performances ($\%$) of the algorithms on the Corel Dataset. Note that the best performance is achieved by the FSG algorithm.}
\label{fig:corel_results}
\end{figure}

In the second set of the experiments, the samples belonging to randomly selected classes are classified. Average (Ave.) and variance (Var.) of the classification performances of the FSG and benchmark algorithms are given in Table \ref{tab:corel_bench}. The classification results given in the tables are depicted in Fig. \ref{fig:corel_results}. 

In the experiments, the difference between the classification performances of the FSG and benchmark algorithms increases as the number of classes ($C$) increases. The performance of the Adaboost algorithm decreases faster than the other algorithms as $C$ increases (see Fig. \ref{fig:corel_results}). Moreover, the Adaboost algorithm performs better than the other benchmark algorithms for classifying the samples belonging to $C \leq 5$ classes. However, the difference between the performance of the Adaboost and the other benchmark algorithms decreases for the classification of samples belonging to $C \geq 5$ classes. Moreover, the performance of the Adaboost and the FSG is approximately same for $C=2$ class classification. Note also that the difference between their performances increases as $C$ increases. In addition, $1$-NN classifier outperforms the Adaboost and is competitive to the other benchmark classifiers for $C \geq 7$.

\begin{table}[htbp]
  \centering
  \caption{Classification results of the algorithms on the Corel Dataset.}
    \begin{tabular}{ccccccccccc}
    \toprule
   $C$ & \multicolumn{2}{c}{\textbf{ Adaboost}} & \multicolumn{2}{c}{\textbf{Rotation Forest}} & \multicolumn{2}{c}{\textbf{Random Subspace}} & \multicolumn{2}{c}{\textbf{1 NN}} & \multicolumn{2}{c}{\textbf{FSG}}\\
          & \textbf{Ave.} & \textbf{Var.} & \textbf{Ave.} & \textbf{Var.} & \textbf{Ave.} & \textbf{Var.} & \textbf{Ave.} & \textbf{Var.} & \textbf{Ave.} & \textbf{Var.}\\
    2     & 90.56\% & 9.30\% & 86.00\% & 0.97\% & 88.11\% & 0.75\% & 82.44\% & 2.78\% & 91.00\% & 0.43\% \\
    3     & 81.33\% & 0.97\% & 76.27\% & 0.57\% & 75.87\% & 0.62\% & 75.27\% & 0.55\% & 86.97\% & 0.53\% \\
    4     & 73.45\% & 0.54\% & 69.75\% & 0.81\% & 70.45\% & 1.27\% & 69.60\% & 1.10\% & 83.85\% & 0.59\% \\
    5     & 64.32\% & 0.32\% & 62.72\% & 0.78\% & 65.32\% & 0.92\% & 61.08\% & 0.65\% & 74.32\% & 0.42\% \\
    6     & 61.17\% & 0.86\% & 61.67\% & 0.83\% & 64.20\% & 1.24\% & 60.50\% & 0.84\% & 71.90\% & 0.67\% \\
    7     & 54.12\% & 0.67\% & 58.00\% & 0.51\% & 62.98\% & 0.45\% & 56.98\% & 0.55\% & 68.65\% & 0.44\% \\
    8     & 53.17\% & 0.12\% & 60.03\% & 0.30\% & 54.92\% & 2.36\% & 58.22\% & 0.35\% & 68.72\% & 0.28\% \\
    9     & 49.02\% & 1.35\% & 56.98\% & 1.81\% & 55.89\% & 3.37\% & 54.98\% & 1.87\% & 67.82\% & 1.16\% \\
    10    & 39.65\% & 0.65\% & 48.35\% & 0.27\% & 47.00\% & 0.35\% & 47.60\% & 0.58\% & 59.80\% & 0.37\% \\
    12    & 38.64\% & 0.65\% & 45.57\% & 0.87\% & 43.22\% & 1.13\% & 45.02\% & 0.86\% & 57.46\% & 0.48\% \\
    14    & 33.16\% & 0.66\% & 47.16\% & 0.63\% & 46.81\% & 0.71\% & 45.76\% & 0.85\% & 57.87\% & 0.75\% \\
    16    & 29.54\% & 0.17\% & 40.42\% & 0.24\% & 41.53\% & 0.29\% & 39.86\% & 0.31\% & 52.07\% & 0.44\% \\
    18    & 25.30\% & 0.59\% & 41.56\% & 0.42\% & 40.91\% & 0.47\% & 39.97\% & 0.44\% & 51.09\% & 0.47\% \\
    20    & 19.46\% & 0.14\% & 38.27\% & 0.16\% & 39.98\% & 0.21\% & 36.25\% & 0.24\% & 47.77\% & 0.20\% \\
    25    & 16.15\% & 0.23\% & 35.92\% & 0.42\% & 35.57\% & 0.63\% & 33.94\% & 0.37\% & 45.84\% & 0.42\% \\
    30    & 14.37\% & 0.55\% & 33.53\% & 0.22\% & 36.28\% & 0.58\% & 32.43\% & 0.26\% & 41.33\% & 0.52\% \\
    \bottomrule
    \end{tabular}%
  \label{tab:corel_bench}%
\end{table}%

\newpage
\textbf{7.2.2.2 Experiments on Caltech Dataset}

In Caltech Dataset experiments, the samples belonging to $2$ to $10$ different classes are randomly selected from the dataset. The experiments are repeated $10$ times for each class. 

In the experiments, the features provided by Gehler and Nowozin \cite{caltech} are used for the construction of the feature spaces. Four feature spaces are constructed using three visual descriptors. Two features spaces consist of SIFT features extracted on a gray scale and an $\mathcal{H} \mathcal{S} \mathcal{I}$ image. The third and the fourth feature spaces contain the features extracted using Region Covariance and Local Binary Patterns descriptors. Implementation details of the features are given below \cite{caltech}.

\begin{table}[htbp]
  \centering
  \caption{Classification results of the benchmark algorithms on the Caltech Dataset.}
    \begin{tabular}{ccccccccccc}
    \toprule
      & \multicolumn{10}{c}{\textbf{Benchmark Algorithms}} \\
    \midrule
    $C$      & \multicolumn{2}{c}{\textbf{ Adaboost}} & \multicolumn{2}{c}{\textbf{Rotation Forest}} & \multicolumn{2}{c}{\textbf{Random Subspace}} & \multicolumn{2}{c}{\textbf{1 NN}} & \multicolumn{2}{c}{\textbf{FSG}} \\
          & \textbf{Ave.} & \textbf{Var. } & \textbf{Ave.} & \textbf{Var. } & \textbf{Ave.} & \textbf{Var. } & \textbf{Ave.} & \textbf{Var. } & \textbf{Ave.} & \textbf{Var. } \\
    2     & 96.47\% & 0.13\% & 87.72\% & 2.86\% & 87.70\% & 1.31\% & 87.78\% & 2.00\% & 95.64\% & 0.28\% \\
    3     & 89.68\% & 0.11\% & 80.90\% & 0.46\% & 81.20\% & 0.33\% & 80.90\% & 0.46\% & 90.46\% & 0.12\% \\
    4     & 81.21\% & 1.55\% & 74.17\% & 1.82\% & 76.10\% & 1.73\% & 72.20\% & 2.62\% & 85.32\% & 0.70\% \\
    5     & 83.27\% & 0.95\% & 77.66\% & 0.92\% & 76.91\% & 1.07\% & 77.55\% & 1.24\% & 88.57\% & 0.41\% \\
    6     & 85.14\% & 0.69\% & 82.73\% & 0.47\% & 83.42\% & 0.51\% & 80.97\% & 0.97\% & 92.15\% & 0.25\% \\
    7     & 77.00\% & 0.55\% & 76.86\% & 0.32\% & 76.79\% & 0.49\% & 76.71\% & 0.25\% & 88.54\% & 0.23\% \\
    8     & 68.49\% & 1.14\% & 71.46\% & 0.97\% & 70.13\% & 1.07\% & 66.77\% & 2.83\% & 85.89\% & 0.35\% \\
    9     & 75.48\% & 0.88\% & 75.90\% & 0.71\% & 75.93\% & 0.83\% & 75.69\% & 0.76\% & 86.28\% & 0.24\% \\ 
    10    & 64.30\% & 0.34\% & 65.66\% & 0.20\% & 65.47\% &	0.18\% & 62.30\% & 0.30\% & 81.06\% & 0.23\% \\
    \bottomrule
    \end{tabular}%
  \label{tab:caltec_bench}%
\end{table}%

In the experiments with Caltech dataset, classification performances of the algorithms do not decrease linearly by increasing number of classes as observed in the experiment with Corel dataset. Note that this non-linear performance variation is observed for all of the aforementioned algorithms. This may be occurred because of the uncertainty in the feature vectors of the samples which is caused by the descriptors employed on the dataset instead of the instability of the classification algorithms. For instance, Adaboost performs better than the other algorithms for $C=2$. In addition, the difference between the classification performances of Adaboost and FSG is $0.78 \%$ for $C=3$. However, the difference increases to $17.40 \%$ for $C=8$ where Adaboost performs worse than the other algorithms.

\subsection{Experiments for Multi-modal Target Detection and Recognition}

Sensors with multiple modalities have the capability of sensing the environment by evaluating the data which represent the different characteristics of the environment. Therefore, the manipulation and the integration of different type of sensors by Decision Fusion algorithms is an important obstacle for various research fields such as robotics. One of the challenging problems of Multi-modal Target Detection and Recognition is to select the \textit{best} information extractors (e.g. feature extractors and classifiers) that manipulate on the multi-modal data which are obtained from different sensors, and achieve inference from the data by reducing the sensor inaccuracy and the environment uncertainty, thereby, the entropy of the data representing the environment. Since the physical modality of each group of data obtained from each individual sensor is discrete and divergent, individual information extractors which are expert on each modality are required. Moreover, the information extractors should be complementary in order to supply the decisions made by each individual expert. 

Decision Fusion algorithms, which employ ensemble learning approach such as Adaboost, often process the data sampled from the same distribution and they are experienced with overfitting of the data as observed in the experiments given in the previous sections. Therefore, most of the Decision Fusion systems may not satisfy the requirements of multi-modal sensor fusion such as the manipulation of the heterogeneous data by considering the classification or generalization error minimization criteria. In order to meet these requirements, FSG is implemented for target detection and recognition using classification in this subsection. The object detection problem is considered as a multi-class classification problem, where background and each target of interest belong to individual classes.

\begin{figure}[ht]
\centering
\vspace{-2mm}
\includegraphics[width=5in, height=4in,angle=0]{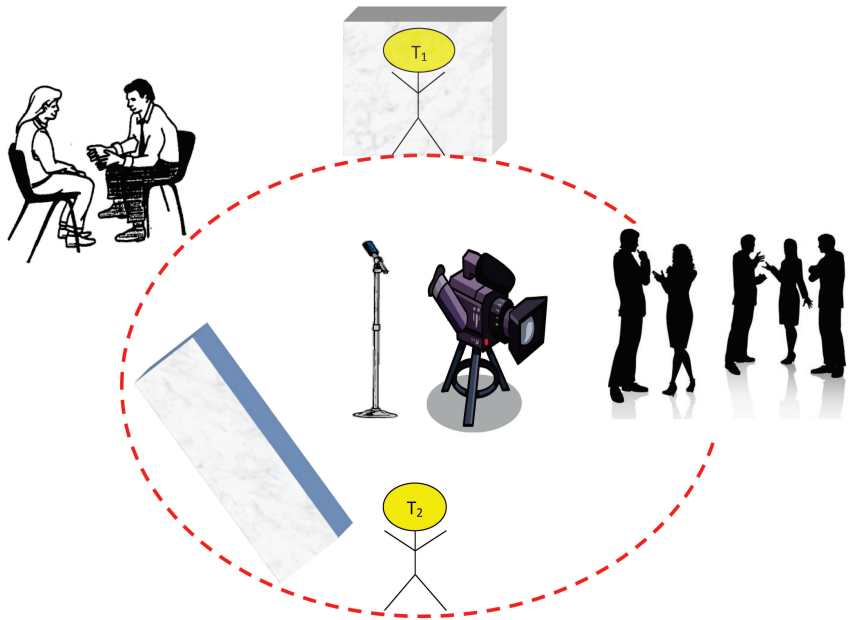}
\caption{The data acquisition setup for the multi-modal decision fusion. }
\label{fig:data_acquisition}
\end{figure}

In the multi-modal target detection and recognition problem, data acquisition is accomplished by an audio-visual sensor, which is a webcam with a microphone located in an indoor environment as shown in Fig. \ref{fig:data_acquisition}. In this scenario, recordings of the audio and video data are obtained from randomly moving two targets $T_{1}$ and $T_{2}$, i.e. two randomly walking people, in the indoor environment. The problem is defined as the classification of the audio and video frames with two targets in the noisy environment, where the other people talking in the environment and the obstacles distributed in the room are the sources of the noise for audio and video data.  

Four classes are defined for the dataset. The first class represents the absence of the moving targets, in other words, there is no target in the environment. The second and the third classes represent the existence of the first and the second target in the environment. In the fourth class, both of the targets take place in the environment. Definitions of the classes according to the presence and absence of two targets $T_1$ and $T_2$ in the environment are given in Table \ref{tab:target_class}. 

\begin{table}[htbp]
  \centering
  \caption{Definitions of classes, according to the presence ($\bigstar$) and absence ($\bigcirc$) of two targets, $T_1$ and $T_2$, in the environment at the same time.}
    \begin{tabular}{cccccc}
    \toprule
          & Class1 & Class2 & Class3 & Class4  \\
    \midrule
    $T_1$ & $\bigcirc$   & $\bigstar$   & $\bigcirc$   & $\bigstar$    \\
    $T_2$  & $\bigcirc$   & $\bigcirc$   & $\bigstar$   & $\bigstar$   \\
    \bottomrule
    \end{tabular}%
  \label{tab:target_class}%
\end{table}%

\begin{table}[htbp]
  \centering
  \caption{Number of samples.}
    \begin{tabular}{cccccc}
    \toprule
          & Class1 & Class2 & Class3 & Class4 & Total \\
    \midrule
    Train & 190   & 190   & 190   & 189   & 759 \\
    Test  & 190   & 190   & 160   & 189   & 729 \\
    \bottomrule
    \end{tabular}%
  \label{tab:multimodal_samples}%
\end{table}%

The audio characteristics of the targets are determined with specific musical melodies with different tonalities. In Table \ref{tab:multimodal_samples}, the number of samples (image frames) belonging to each class for each data set is given.

\begin{figure}[ht]
\centering
\vspace{-2mm}
\includegraphics[width=5in, height=3in,]{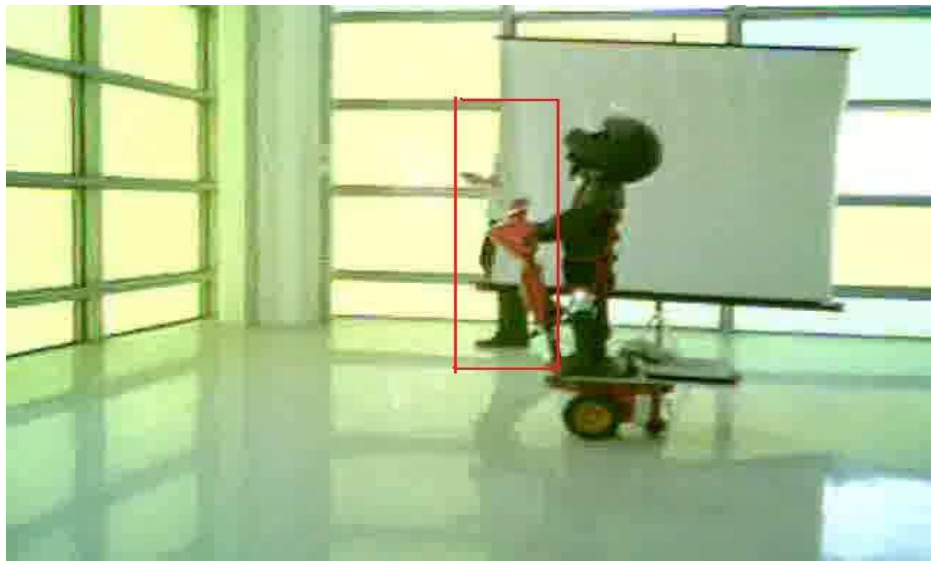}
\caption{A sample frame used in the training dataset in which a target ($T_1$) is hidden behind an obstacle which is a curtain. }
\label{fig:data_target1}
\end{figure}

The experimental setup is designed to achieve complementary expertise of the base-layer classifiers on different classes. For instance, if a target is hidden behind an obstacle such as a curtain (see Fig. \ref{fig:data_target1}), then a base-layer classifier which employs audio features for classification can correctly detect the target behind the curtain, even if a base-layer classifier which employs visual features for classification, cannot detect the target correctly.

In the experiments, two MPEG-7 descriptors, Homogenous Texture (HT) and Color Layout (CL), and three audio descriptors, Fluctuation (Fluct.), Chromagram (Chrom.) and Mel-Frequency Cepstral Coefficients (MFCC), \cite{mfcc} are used to extract visual and audio features, respectively \cite{mfcc}. FSG is used for the fusion of the decisions of the classifiers employed on $i$) visual features ( \textit{\textbf{Video Fusion}} ), $ii$) audio features ( \textit{\textbf{Audio Fusion}} ) and $iii$) both audio and visual features ( \textit{\textbf{Audio-Visual Fusion}} ).
 
Experimental results show that the base-layer classifiers employed on visual features perform better than the classifiers employed on audio features for the fourth class. However, the classifiers employed on audio features perform better than the classifiers employed on visual features for the first three classes. For instance, the base-layer classifiers employed on the visual descriptors most likely misclassify the samples from the second class, but perform better than the other classifiers for the fourth class (see Table \ref{tab:multimodal_training} and Table \ref{tab:multimodal_test}). On the other hand, the base-layer classifiers employed on audio descriptors have a better discriminative power compared to the base-layer classifiers employed on the visual descriptors for the first class.

One of the reasons of this observation is that the classifiers employing audio features, which are affected by audio noise, are less sensitive to \textit{noise} than the classifiers employed on visual features which are affected by visual \textit{noise}. In other words, two targets have visual appearance properties similar to the other objects in the environment, and the obstacles (e.g. curtains and doors) block completely the visual appearance of the targets. On the other hand, the targets have different visual appearance properties such that the heights of the targets and color of their clothes are different from each other. In addition, the audio properties of the measurements obtained from the targets have discriminative characteristics which are different than the other objects in the environment. 

An analysis of Table \ref{tab:multimodal_training} and Table \ref{tab:multimodal_test} reveals that the performance of an individual descriptor varies across the classes due to similar arguments. As a result, a substantial increase in the general classification performance of the FSG is achieved. 

\begin{table}[htbp]
  \centering
  \caption{Classification performances for training dataset.}
    \begin{tabular}{rccccc}
    \toprule
          & Class1 & Class2 & Class3 & Class4 & Total \\
    \midrule
    \multicolumn{1}{c}{Homogeneous Texture} & 76.84\% & 67.89\% & 76.84\% & 96.30\% & 79.45\% \\
    \multicolumn{1}{c}{Color Layout} & 93.16\% & 86.84\% & 84.21\% & \textbf{97.35\%} & 90.38\% \\
    \multicolumn{1}{c}{MFCC} & \textbf{99.47\%} & 84.74\% & \textbf{94.74\%} & 83.60\% & \textbf{90.65\%} \\
    \multicolumn{1}{c}{Chromagram} & 98.42\% & \textbf{90.00\%} & 89.47\% & 82.01\% & 89.99\% \\
    \multicolumn{1}{c}{Fluctuation} & 94.74\% & 85.79\% & 75.79\% & 52.38\% & 77.21\% \\

  \midrule  
    \multicolumn{1}{c}{\textit{\textbf{Video Fusion}}} & 92.63\% & 87.37\% & 84.21\% & 95.77\% & 89.99\% \\
    \multicolumn{1}{c}{\textit{\textbf{Audio Fusion}}} & 97.89\% & 93.16\% & 96.32\% & 92.59\% & 94.99\% \\
    \multicolumn{1}{c}{\textit{\textbf{Audio-Visual Fusion}}} & 99.47\% & 97.89\% & 98.42\% & 100.00\% & 98.95\% \\
    \bottomrule
    \end{tabular}%
  \label{tab:multimodal_training}%
\end{table}%

\begin{table}[htbp]
  \centering
  \caption{Classification performances for test dataset.}
    \begin{tabular}{rccccc}
    \toprule
          & Class1 & Class2 & Class3 & Class4 & Total \\
    \midrule
    \multicolumn{1}{c}{Homogeneous Texture} & 54.74\% & 49.47\% & 43.75\% & \textbf{93.12\%} & 60.91\% \\
    \multicolumn{1}{c}{Color Layout} & 76.32\% & 49.47\% & 40.63\% & 83.07\% & 63.24\% \\
    \multicolumn{1}{c}{MFCC} & 92.11\% & 77.37\% & \textbf{93.13\%} & 81.48\% & \textbf{85,73\%} \\
    \multicolumn{1}{c}{Chromagram} & 92.63\% & \textbf{84.21\%} & 83.13\% & 66.67\% & 81.62\% \\
    \multicolumn{1}{c}{Fluctuation} & \textbf{93.68\%} & 82.63\% & 75.00\% & 52.38\% & 75.99\% \\
      \midrule  

    \multicolumn{1}{c}{\textit{\textbf{Video Fusion}}} & 69.47\% & 54.21\% & 45.63\% & 90.48\% & 65.71\% \\
    \multicolumn{1}{c}{\textit{\textbf{Audio Fusion}}} & 90.53\% & 93.16\% & 93.13\% & 79.37\% & 88.89\% \\
    \multicolumn{1}{c}{\textit{\textbf{Audio-Visual Fusion}}} & 93.68\% & 94.21\% & 94.37\% & 97.88\% & 95.06\% \\
    \bottomrule
    \end{tabular}%
  \label{tab:multimodal_test}%
\end{table}%

Each cell of Table \ref{tab:multimodal_training_covariance} and Table \ref{tab:multimodal_test_covariance} represents the number of samples which are misclassified by the classifier for the descriptor in the $i^{th}$ row, and correctly classified by the classifier for the descriptor in the $j^{th}$ column, for the training and test datasets, respectively. In the tables, the maximum number of misclassified samples for each descriptor is bolded.

For example, $144$ samples which are misclassified in HT feature space are correctly classified in Chromagram feature space. The samples that are misclassified in the feature spaces defined by the visual descriptors are correctly classified in the feature spaces defined by the audio descriptors. This is observed when the visual appearance of the targets are affected by the visual noise, e.g. the targets are completely blocked by an obstacle, such as a curtain, but their sounds are clearly recorded by the audio sensor, as shown in Fig. \ref{fig:data_target1}. Therefore, it can be easily observed from the tables that the feature spaces are complementary to each other. 

On the other hand, the samples that are misclassified in the feature spaces defined by the audio descriptors (e.g. Fluctuation and Chromagram) are correctly classified in the feature spaces defined by the visual descriptors (e.g. CL and HT) when there are other objects that make sounds with audio characteristics similar to the targets in the environment. In this case, audio features of the targets are affected by audio noise. If the visual sensor can make \textit{clear} measurements on the targets, such that the visual features are not affected by visual noise, then the classifiers employed in the feature spaces defined by the visual descriptors can correctly classify the samples.

\begin{table}[htbp]
  \centering
  \caption{Covariance matrix for the number of correctly and misclassified samples in training dataset.}
    \begin{tabular}{cccccccc}
    \toprule
    \multicolumn{2}{c}{Train Dataset} & \multicolumn{5}{c}{\textbf{Correct Classification}} &  \\
    \midrule
    \multirow{7}{*}{\textbf{Misclassification}} &       & HT    & CL    & MFCC  & Chrom. & Fluct. & Total  \\
          & HT    & 0     & 137   & 142   & \textbf{144} & 130   & 156 \\
          & CL    & 54    & 0     & \textbf{64} & 59    & 57    & 73 \\
          & MFCC  & 57    & \textbf{62} & 0     & 44    & 40    & 71 \\
          & Chromagram & \textbf{64} & 62    & 49    & 0     & 39    & 76 \\
          & Fluctuation & 147   & \textbf{157} & 142   & 136   & 0     & 173 \\
          & & & & & & \\
    \bottomrule
    \end{tabular}%
  \label{tab:multimodal_training_covariance}%
\end{table}%

\begin{table}[htbp]
  \centering
  \caption{Covariance matrix for the number of correctly and misclassified samples in test dataset.}
    \begin{tabular}{cccccccc}
    \toprule
    \multicolumn{2}{c}{Test Dataset} & \multicolumn{5}{c}{\textbf{Correct Classification}} &  \\
    \midrule
    \multirow{7}{*}{\textbf{Misclassification}} &       & HT    & CL    & MFCC  & Chrom. & Fluct. & Total \\
          & HT    & 0     & 134   & 247   & \textbf{249} & 233   & 285 \\
          & CL    & 117   & 0     & \textbf{235} & 223   & 216   & 268 \\
          & MFCC  & 66    & \textbf{71} & 0     & 52    & 54    & 104 \\
          & Chromagram & \textbf{98} & 89    & 82    & 0     & 61    & 134 \\
          & Fluctuation & \textbf{123} & \textbf{123} & \textbf{125} & 102   & 0     & 175 \\
          & & & & & & \\
    \bottomrule
    \end{tabular}%
  \label{tab:multimodal_test_covariance}%
\end{table}%


\subsubsection{Statistical Analysis of Feature, Decision and Fusion Spaces on Multi-modal Dataset}

In this subsection, transformations of class conditional distributions of feature and decision vectors through the layers of the FSG in feature, decision and fusion spaces are analyzed for multi-modal target detection and recognition problem. Histograms are used to approximate the distributions for visualization. 

In the histogram representation \cite{hist_ent}, first the range of random variables $X$ that reside in $[0,1]$ (e.g., posterior probabilities computed by base-layer classifiers in the FSG), is divided into $B$ intervals $(low_b,up_b)$, $b=1,2,\ldots,B$ representing the width $w_b=up_b-low_b$ of the $b^{th}$ \textit{bin} of a histogram. Denoting the probability density function of $X$ as $f$, the entropy can be approximated using
\[
\mathbb{H}(X)\approx - \sum \limits _{b=1} ^B p_b,
\]
where the probability of a bin 
\[
p_b=\int _{low_b} ^{up_b}  f(x) dx, \; \forall b=1,2,\ldots,B
\]
is approximated as the area of a rectangle of height $f(x^b)$ which is $w_bf(x^b)$ where $x^b$ is a representative value within the interval $(low_b,up_b)$. Then the entropy is approximated as
\[
\mathbb{H}(X) \approx - \sum \limits _{b=1} ^B p_b \log \frac{p_b}{w_b}.
\]

In order to represent the class conditional distributions using histograms, first the number of samples belonging to each class is computed by counting $n_{b,c}$ which is the number of samples that fall into each of the disjoint histogram bins $b=1,2,\ldots,B$ for each class $c=1,2,\ldots,C$, as 
\[
N_c=\sum \limits _{b=1} ^B n_{b,c}
\] 
and $N=\sum \limits _{c=1} ^C N_{c}$ where $N$ is the number of samples in the dataset. Then a bin probability is computed as $p_{b,c}=\frac{n_{b,c}}{N}$. 

Note that, three different spaces are constructed through the FSG; $i$) feature space (at the input of base-layer classifiers), $ii$) decision space (at the output of base-layer classifiers) and $iii$) fusion space (at the input of meta-layer classifier). Feature spaces consist of the feature sets obtained from the descriptors. A bin probability $p_{b,c}$ of a histogram computed in a feature space $F_j$ is an approximation to a class conditional distribution. In a decision space of a classifier employed on $F_j$, posterior probability or class membership vectors $\bar{\mu} (\bar{x}_{ij})$ are used. In the fusion space, the histograms are computed using the concatenated membership vectors $\bar{\mu}(\bar{x}_{i})$.

\begin{figure}[htbp]
\centering
\subfloat[]{\includegraphics[width=3.5in,height=2.5in]{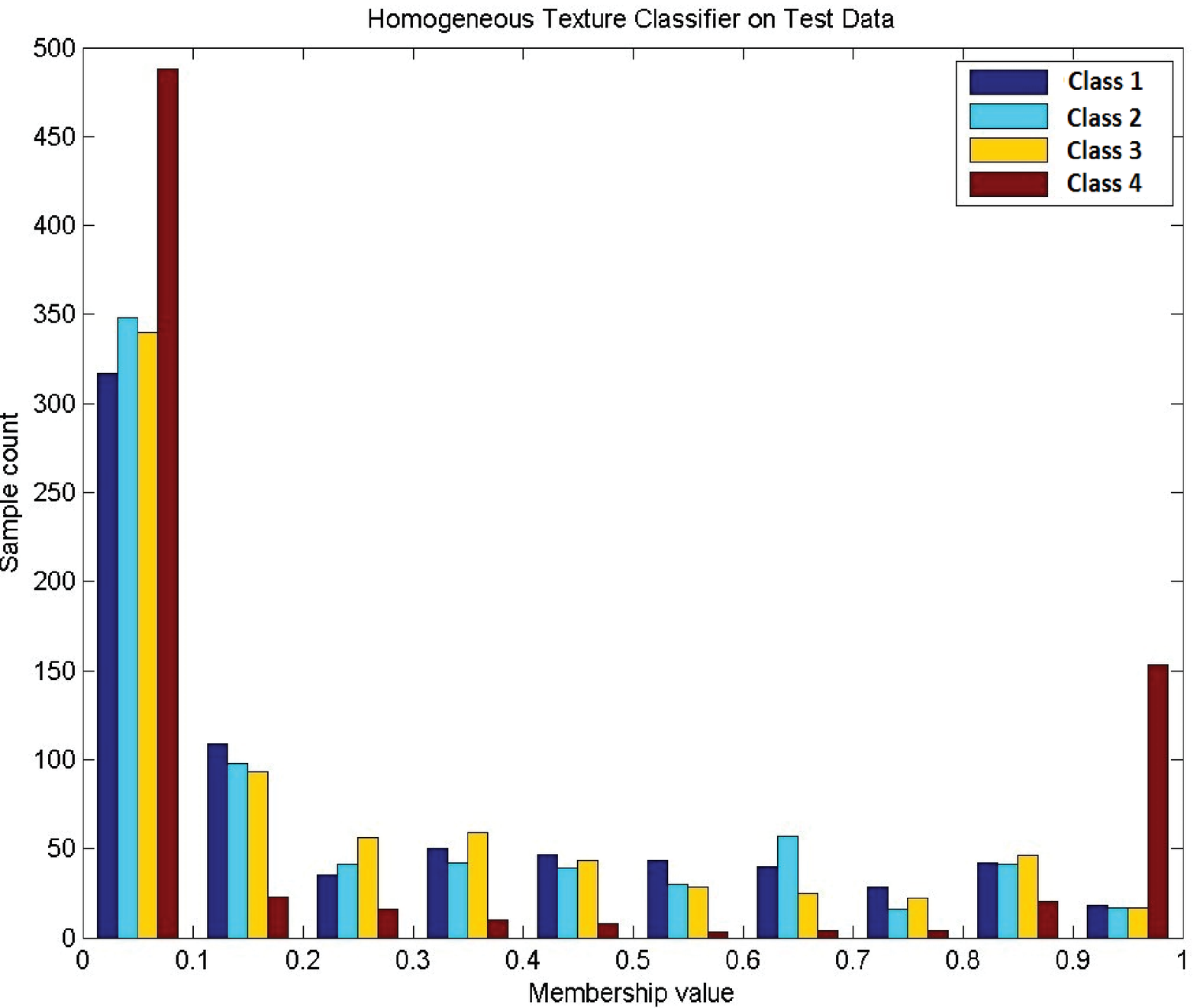}}
\subfloat[]{\includegraphics[width=3.5in,height=2.5in]{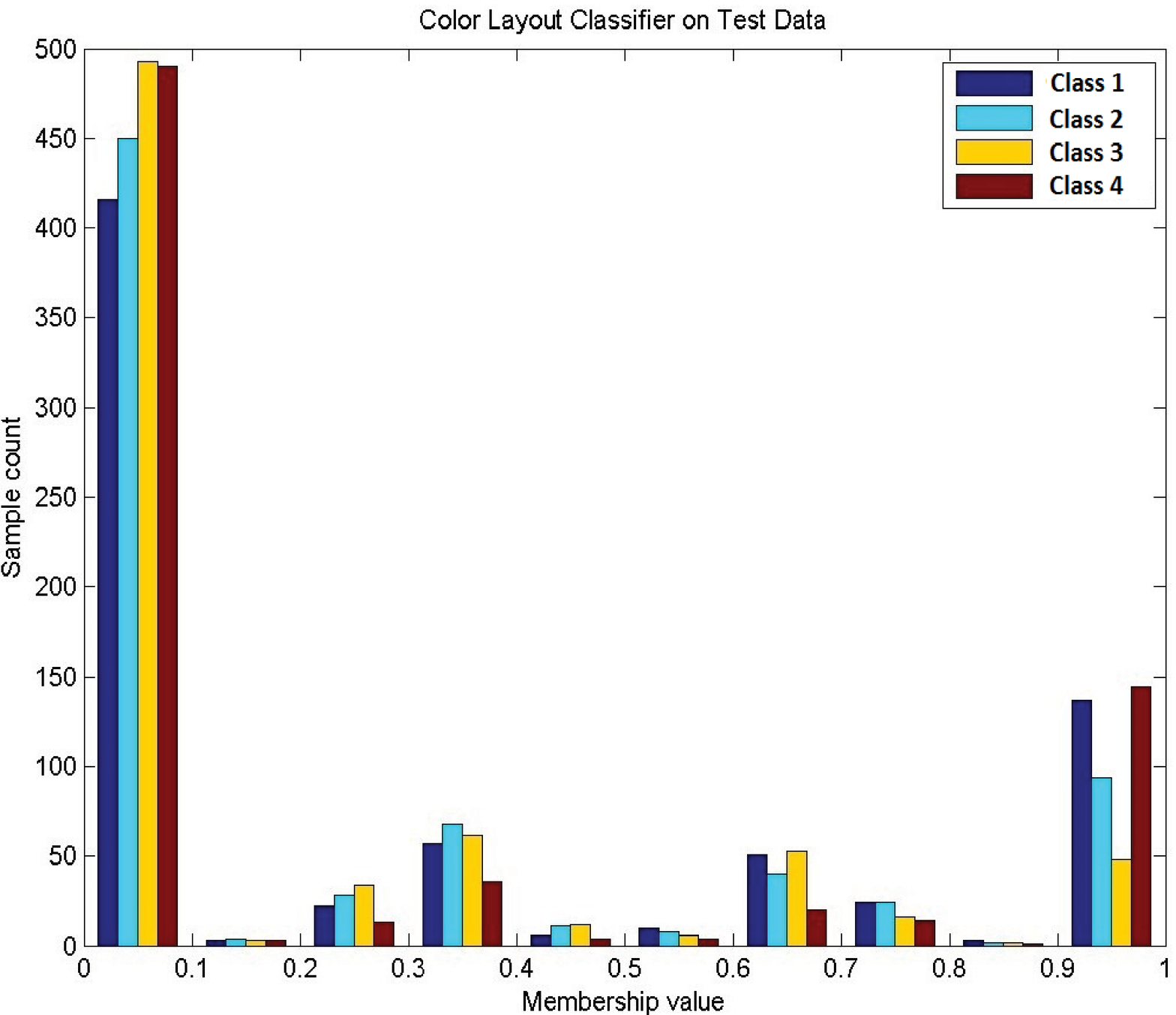}} \\
\subfloat[]{\includegraphics[width=3.5in,height=2.5in]{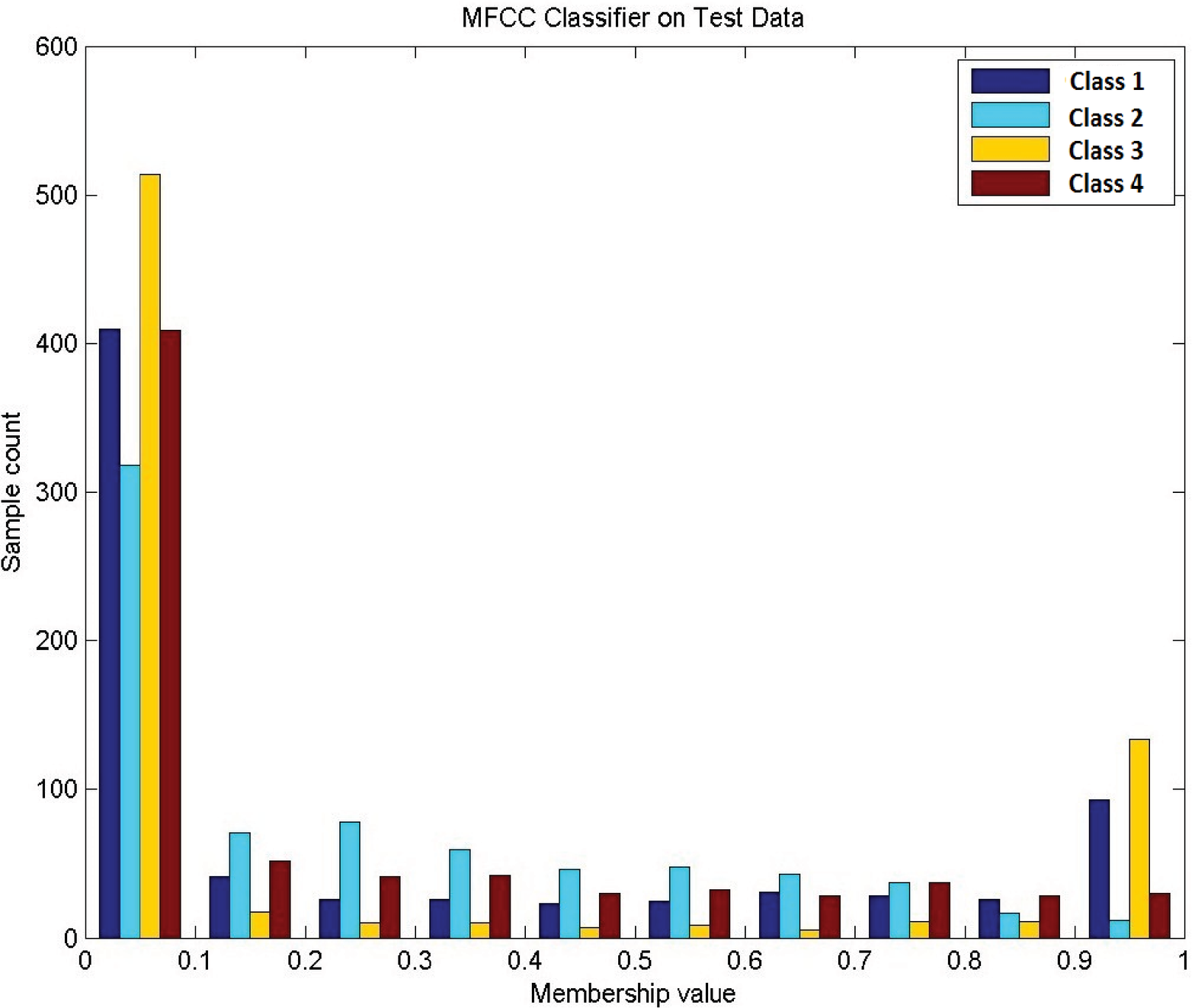}}
\subfloat[]{\includegraphics[width=3.5in,height=2.5in]{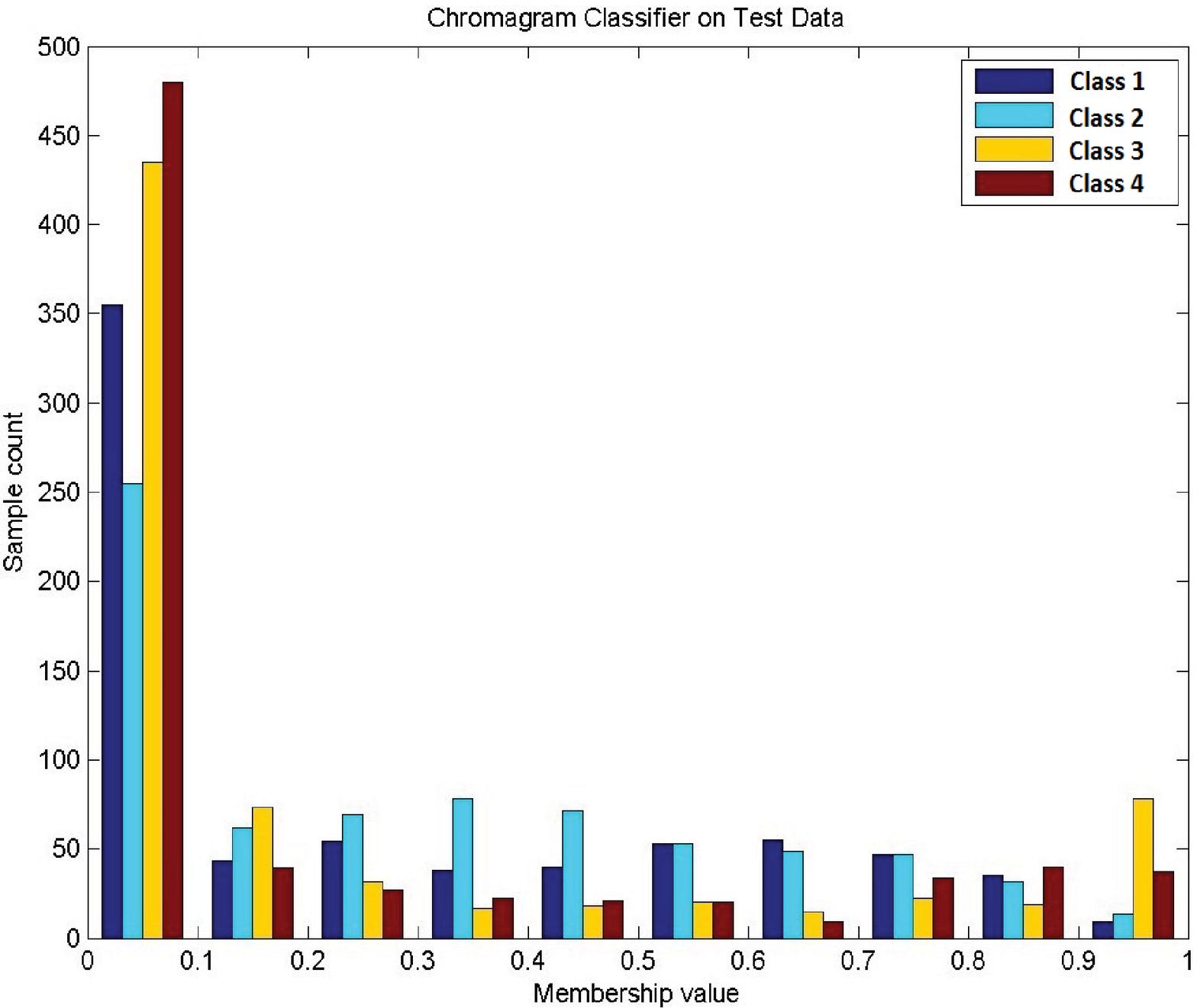}}  \\
\subfloat[]{\includegraphics[width=3.5in,height=2.5in]{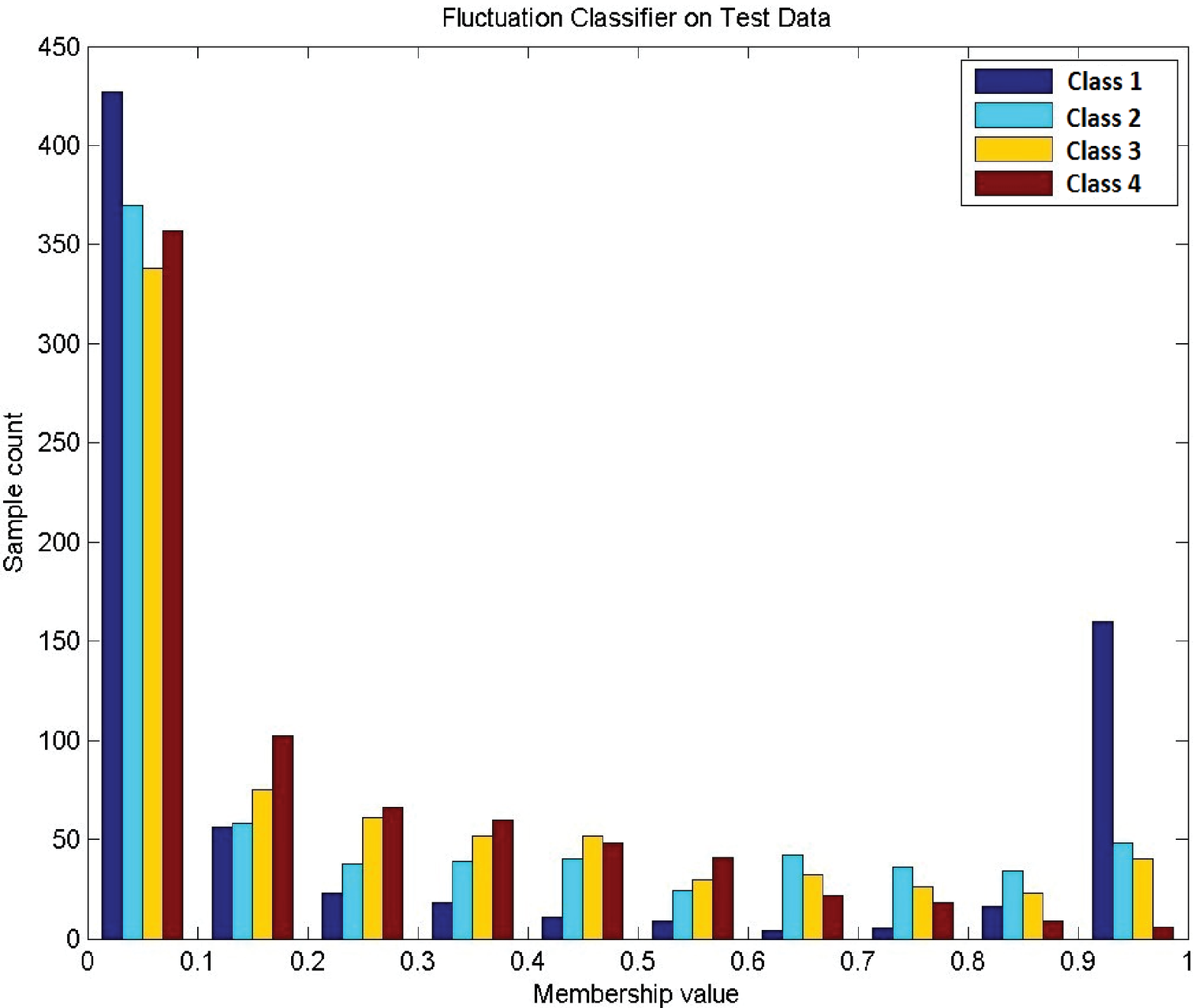}}
\subfloat[]{\includegraphics[width=3.5in,height=2.5in]{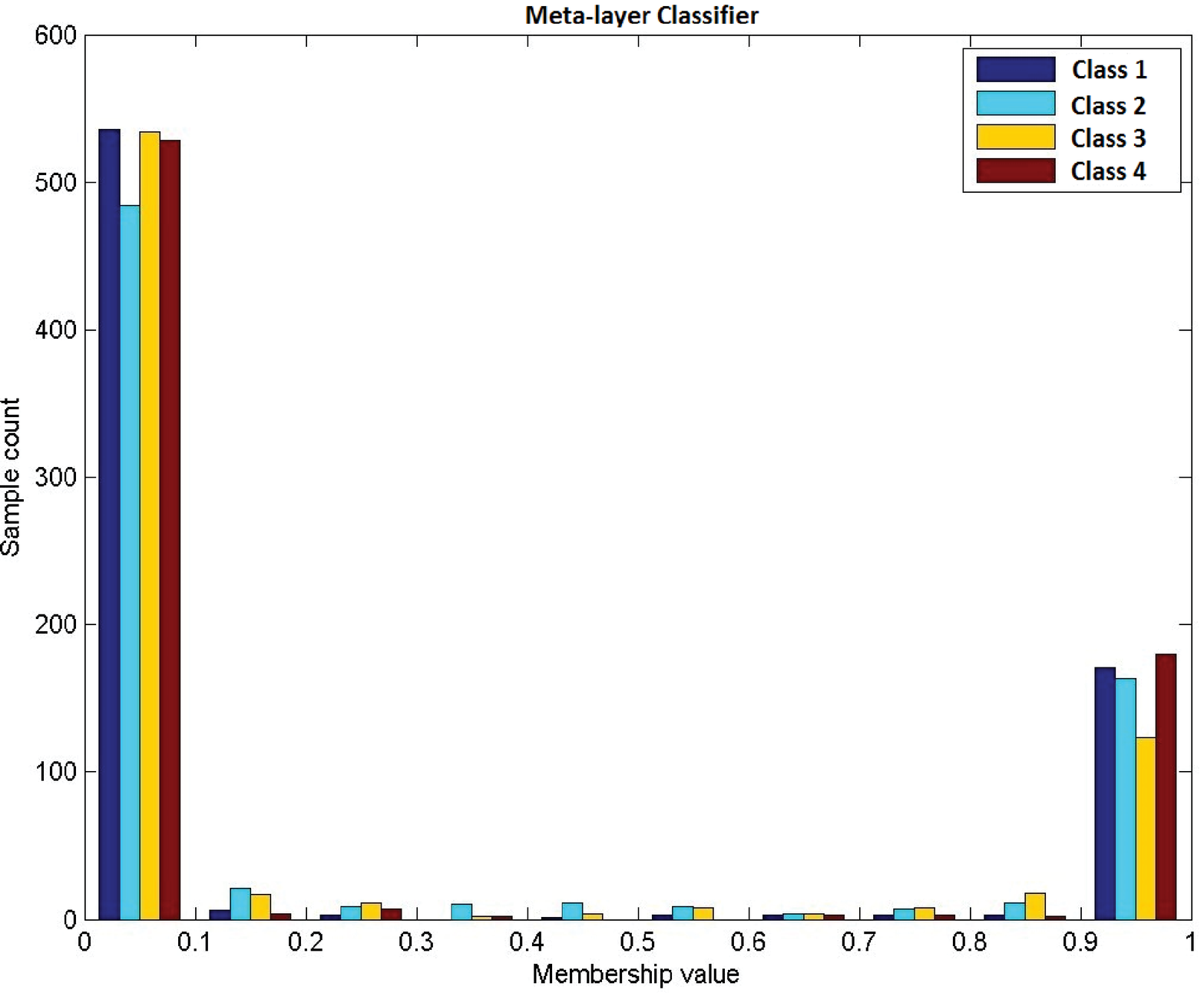}}  \\
\caption{Histograms which represent distributions for the individual decision spaces of base-layer classifiers employed using (a) Histogram Texture, (b) Color Layout, (c) MFCC, (d) Chromagram, (e) Fluctuation features, and (f) in the fusion space of the meta-classifier in FSG. Notice that the lowest entropy is observed in the fusion space.}
\label{fig:entropies}
\end{figure}

In Fig. \ref{fig:entropies}, the histograms representing the approximate probability distributions at each base-layer decision space (Fig. \ref{fig:entropies} (a-e)) and the fusion space (i.e. the meta-layer input feature space) (Fig. \ref{fig:entropies}.f) are displayed for test dataset. It is observed from the histograms that the concatenation operation decreases the uncertainty of the feature spaces.

Entropy values computed in feature spaces are given in Table \ref{tab:multimodal_base_entropy}. Entropy values computed for each class in each feature space give the information about the data uncertainity in the feature space. If the distributions of the features in a feature space $F_j$ provide lower entropy values for a particular class $\omega_c$ than the other classes, then the features may represent a characteristic of class $\omega_c$. Therefore, a classifier employed on $F_j$ classifies the samples belonging to $\omega_c$ with better performance than the samples belonging to other classes.

For instance, distributions of  Fluctuation, MFCC and Homogeneous Texture features have the lowest entropy values for the first, the third and the fourth classes, respectively (see Table \ref{tab:multimodal_base_entropy}). The base-layer classifiers which use these features provide the highest classification performances, as shown in Table \ref{tab:multimodal_test}. 

Although the distribution of Color Layout features gives the lowest entropy for the second class than the other audio and visual features, a base-layer classifier employed on Color Layout features performs worse than the other classifiers employed on other features. However, the  features of the samples belonging to the fourth class have the lowest entropy in Color Layout feature space (see the row of Table \ref{tab:multimodal_base_entropy} labeled Color Layout). Then, a classifier employed on Color Layout feature space gives the highest classification performance for the fourth class  as given in Table \ref{tab:multimodal_test}.

Entropy values computed in decision and fusion spaces are given in Table \ref{tab:multimodal_meta_entropy} for test dataset. Entropy values of the class membership vectors in decision spaces represent the decision uncertainty of base-layer classifiers for each class. Note that the classifiers employed on the feature spaces with minimum decision uncertainties for particular classes provide the highest classification performances for these classes (see Table \ref{tab:multimodal_test}). 

Entropy values of the membership vectors $\bar{\mu}(\bar{x}_{i})$ that reside in the fusion space represent the joint entropy of $ \{ \bar{\mu }(\bar{x}_{i,j} ) \} _{j=1} ^J$, since $\bar{\mu }(\bar{x}_{i} ) = \left[\bar{\mu }(\bar{x}_{i,1} ) \ldots \bar{\mu }(\bar{x}_{i,j} ) \ldots \bar{\mu }(\bar{x}_{i,J} ) \right]$. If classifier decisions are independent, then the entropy value $Ent_{fusion}$ of $\bar{\mu}(\bar{x}_{i})$ is equal to the sum of the entropy values $Ent_{j}$ of $ \bar{\mu }(\bar{x}_{i,j} )$, $\forall j=1,2,\ldots,J$, such that 
\[
Ent_{fusion} = \sum \limits _{j=1} ^J Ent_{j}. 
\]

However, $Ent_{fusion} \leq \sum \limits _{j=1} ^J Ent_{j}$ in Table \ref{tab:multimodal_meta_entropy}, which implies that the decisions are dependent. This dependency occurs by sharing the samples among the classifiers in the FSG as shown in Table \ref{tab:multimodal_test_covariance}. Thereby, lower entropy values are obtained in the fusion space.

\begin{table}[htbp]
  \centering
  \caption{Entropy values computed in feature spaces for test dataset.}
    \begin{tabular}{ccccc}
    \toprule
    Feature Spaces & Class 1 & Class 2 & Class 3 & Class 4 \\
    \midrule
    Homogeneous Texture    & 0.3751 & 0.3840 & 0.3702 & \textbf{0.0679} \\
    Color Layout    & 0.1905 & \textbf{0.2644} & 0.3255 & 0.0861 \\
    MFCC  & 0.1920 & 0.3824 & \textbf{0.0879} & 0.3347 \\
    Chromagram  & 0.3442 & 0.3621 & 0.2011 & 0.2834 \\
    Fluctuation  & \textbf{0.0389} & 0.3013 & 0.3115 & 0.4276 \\
    \bottomrule
    \end{tabular}%
  \label{tab:multimodal_base_entropy}%
\end{table}%

\begin{table}[htbp]
  \centering
  \caption{Entropy values computed in decision and fusion spaces for test dataset.}
    \begin{tabular}{ccccc}
    \toprule
    Decision and Fusion Spaces & Class 1 & Class 2 & Class 3 & Class 4 \\
    \midrule
    Homogeneous Texture    & 0.2160 & 0.2360 & 0.2550 & \textbf{0.0457} \\
    Color Layout    & 0.1057 & 0.3052 & 0.2383 & 0.4584 \\
    MFCC  & 0.1539 & 0.2161 & \textbf{0.1322} & 0.1936 \\
    Chromagram  & 0.1165 & \textbf{0.1092} & 0.1582 & 0.1760 \\
    Fluctuation  & \textbf{0.0344} & 0.2286 & 0.2890 & 0.3228 \\
    Fusion Space & \textit{0.0228} & \textit{0.0529} & \textit{0.0873} & \textit{0.0156} \\
    \bottomrule
    \end{tabular}%
  \label{tab:multimodal_meta_entropy}%
\end{table}%

\section{Summary and Conclusion}
\label{sec:fsg_summary}

In this work, the classification error minimization problem has been addressed for Decision Fusion using an ensemble of $k$- Nearest Neighbor ($k$-NN) classifiers as the minimization of the difference between $N$-sample and large-sample classification errors of the nearest neighbor classifiers. 

$k$-NN algorithm has been employed for the analysis of the error difference minimization problem because of three reasons. First, the error of $k$-NN is upper and lower bounded by the Bayes Error which is the minimum achievable classification error by any classification algorithm. Therefore, the error bounds are tractable. Second, $k$-NN can be considered as a decision fusion algorithm which combines the decisions of the neighbor samples of a given test sample to estimate its label or class membership value. In addition, $k$-NN is a \textit{powerful} nonparametric density estimation algorithm used for the estimation of posterior probabilities to design distance functions. 

Distance learning problem for classification error minimization is analyzed as a feature, decision and fusion space design, i.e. classifier design and decision fusion, problem. In order to solve these problems, a hierarchical decision fusion algorithm called Fuzzy Stacked Generalization (FSG) is employed.

Base-layer classifiers of the FSG are used for two purposes; $i$) mapping feature vectors to decision vectors and $ii$) estimating posterior probabilities, which are the variables of the distance function, using the datasets. Decision vectors in the decision spaces are concatenated to construct the feature vectors in the fusion space for a meta-layer classifier.

One of the major contributions of the suggested decision fusion method is to minimize the difference between $N$-sample and large-sample classification error of $k$-NN. This property is shown by using the distance learning approach of Short and Fukunaga \cite{short}. In addition, samples should be classified with complete certainty as shown by Cover and Hart \cite{cover} in order to converge the large-sample classification error to Bayes Error. In FSG, this condition should be satisfied by the meta-layer classifier employed on the fusion space in almost everywhere, i.e. by each feature vector of each sample. 


The proposed FSG algorithm is tested on artificial and benchmark datasets by the comparisons with the state of the art algorithms such as Adaboost, Rotation Forest and Random Subspace. In the first group of the experiments, the samples belonging to different classes are gradually overlapped in the artificial datasets. The experiments are designed in such a way that the requirements of $N$-sample and large-sample error minimization conditions are controlled. 

It is observed that if one can design the feature spaces such that $\hat{Ave}_{corr} \approx 1$, the classification performance of FSG becomes significantly higher than that of the individual classifier performances. This experiment also shows that the performance of FSG depends on sharing and collaborating the features of the samples rather than the performance of individual classifiers. 

In the second group of experiments, the proposed algorithms are compared with state of the art algorithms using benchmark datasets. In two class multi-attribute classification experiments, FSG and Adaboost provide similar performances. Meanwhile, Adaboost outperforms FSG for the classification of samples with high dimensional features, i.e. large number of attributes such as $A=20$. This is observed due to the curse-of-dimensionality observed in the fusion space which is $20 \times 2=40$ dimensional. 

FSG outperforms state of the art algorithms in the experiments employed on multi-feature datasets, especially for the classification of the samples belonging to $C>2$ number of classes. The difference between the classification performances of the FSG and the state of the art algorithms employed on multi-feature datasets is greater than that of the difference observed in multi-attributed datasets because of two reasons. First, the proposed algorithms fix the dimensions of the feature vectors in fusion space to $CJ$ (number of classes $\times$ number of feature extractors) no matter how high is the dimension $D_j$ of the individual feature vectors at the base-layer. Second, employing distinct feature extractors for each base-layer classifier enables us to split various attributes of the feature spaces, coherently. Therefore, each base-layer classifier gains an expertise to learn a specific property of a sample, and correctly classifies a group of samples belonging to a certain class in the training data. This approach assures the diversity of the classifiers as suggested by Kuncheva \cite{comb_kun} and enables the classifiers to collaborate for learning the classes or groups of samples. It also allows us to optimize the parameters of each individual base-layer classifier independent of the other.  

In the third group of experiments, we constructed a multi-modal dataset using different sensors, namely audio and video recorders. This multi-modal data is fused under the FSG architecture. Apparently, the features extracted from the individual modes have different statistical properties, and give diverse information about different classes. Therefore, base-layer classifiers each of which can correctly classify the samples belonging to specific classes can be trained, even if the individual performances of the classifiers are low. Since this data setting is complementary to the observations on the artificial datasets (see Section \ref{sec:synthetic}), the FSG boosts the performances of the base-layer classifiers with $10\%$ performance gain.

Moreover, it is observed that the entropies of the features are decreased through the feature space transformations from the base-layer to the meta-layer of the architecture. Therefore, the architecture transforms the linearly non-separable feature spaces with higher dimensions into a more separable feature space (Fusion Space) with lower dimensions which are fixed with the number of classes and base-layer classifiers.

\bibliography{IEEEabrv,thesis_references}

\begin{thebibliography}{10}
\providecommand{\url}[1]{#1}
\csname url@samestyle\endcsname
\providecommand{\newblock}{\relax}
\providecommand{\bibinfo}[2]{#2}
\providecommand{\BIBentrySTDinterwordspacing}{\spaceskip=0pt\relax}
\providecommand{\BIBentryALTinterwordstretchfactor}{4}
\providecommand{\BIBentryALTinterwordspacing}{\spaceskip=\fontdimen2\font plus
\BIBentryALTinterwordstretchfactor\fontdimen3\font minus
  \fontdimen4\font\relax}
\providecommand{\BIBforeignlanguage}[2]{{%
\expandafter\ifx\csname l@#1\endcsname\relax
\typeout{** WARNING: IEEEtran.bst: No hyphenation pattern has been}%
\typeout{** loaded for the language `#1'. Using the pattern for}%
\typeout{** the default language instead.}%
\else
\language=\csname l@#1\endcsname
\fi
#2}}
\providecommand{\BIBdecl}{\relax}
\BIBdecl

\bibitem{Wolpert}
D.~Wolpert, ``Stacked generalization,'' \emph{Neural Netw.}, vol.~5, no.~2, pp.
  241--259, 1992.

\bibitem{Ueda}
N.~Ueda, ``Optimal linear combination of neural networks for improving
  classification performance,'' \emph{IEEE Transactions on Pattern Analysis and
  Machine Intelligence}, vol.~22, no.~2, pp. 207--215, Feb 2000.

\bibitem{sen}
M.~U. Sen and H.~Erdogan, ``Linear classifier combination and selection using
  group sparse regularization and hinge loss,'' \emph{Pattern Recognition
  Letters}, vol.~34, no.~3, pp. 265--274, 2013.

\bibitem{Rooney}
N.~Rooney, D.~Patterson, and C.~Nugent, ``Non-strict heterogeneous stacking,''
  \emph{Pattern Recognition Letters}, vol.~28, no.~9, pp. 1050--1061, 2007.

\bibitem{Zenko}
B.~Zenko, L.~Todorovski, and S.~Dzeroski, ``A comparison of stacking with meta
  decision trees to bagging, boosting, and stacking with other methods,'' in
  \emph{Proceedings of the 2001 IEEE International Conference on Data Mining},
  ser. ICDM '01.\hskip 1em plus 0.5em minus 0.4em\relax Washington, DC, USA:
  IEEE Computer Society, 2001, pp. 669--670.

\bibitem{Tan}
X.~Tan, S.~Chen, Z.-H. Zhou, and F.~Zhang, ``Recognizing partially occluded,
  expression variant faces from single training image per person with som and
  soft k-nn ensemble,'' \emph{IEEE Transactions on Neural Networks}, vol.~16,
  no.~4, pp. 875--886, Jul 2005.

\bibitem{Ting}
K.~M. Ting and I.~H. Witten, ``Issues in stacked generalization,''
  \emph{Journal of Artificial Intelligence Research}, vol.~10, no.~1, pp.
  271--289, May 1999.

\bibitem{adaboost}
R.~E. Schapire, ``A brief introduction to boosting,'' in \emph{Proceedings of
  the 16th international joint conference on Artificial intelligence - Volume
  2}, ser. IJCAI'99.\hskip 1em plus 0.5em minus 0.4em\relax San Francisco, CA,
  USA: Morgan Kaufmann Publishers Inc., 1999, pp. 1401--1406.

\bibitem{random_subspace}
T.~K. Ho, ``The random subspace method for constructing decision forests,''
  \emph{IEEE Transactions on Pattern Analysis and Machine Intelligence},
  vol.~20, no.~8, pp. 832--844, Aug 1998.

\bibitem{rf}
J.~J. Rodriguez, L.~I. Kuncheva, and C.~J. Alonso, ``Rotation forest: A new
  classifier ensemble method,'' \emph{IEEE Transactions on Pattern Analysis and
  Machine Intelligence}, vol.~28, no.~10, pp. 1619--1630, Oct 2006.

\bibitem{Ghorbani}
A.~Ghorbani and K.~Owrangh, ``Stacked generalization in neural networks:
  generalization on statistically neutral problems,'' in \emph{IEEE
  International Joint Conference on Neural Networks}, vol.~3, 2001, pp.
  1715--1720.

\bibitem{zhao}
G.~Zhao, Z.~Shen, C.~Miao, and R.~Gay, ``Enhanced extreme learning machine with
  stacked generalization,'' in \emph{IEEE International Joint Conference on
  Neural Networks}, 2008, pp. 1191--1198.

\bibitem{iciar}
M.~Ozay and F.~T. Vural, ``On the performance of stacked generalization
  classifiers,'' in \emph{Proceedings of the 5th international conference on
  Image Analysis and Recognition}, ser. ICIAR '08.\hskip 1em plus 0.5em minus
  0.4em\relax Berlin, Heidelberg: Springer-Verlag, 2008, pp. 445--454.

\bibitem{emre}
E.~Akbas and F.~Yarman~Vural, ``Automatic image annotation by ensemble of
  visual descriptors,'' in \emph{IEEE Conference on Computer Vision and Pattern
  Recognition}, 2007, pp. 1--8.

\bibitem{sigletos}
G.~Sigletos, G.~Paliouras, C.~D. Spyropoulos, and M.~Hatzopoulos, ``Combining
  information extraction systems using voting and stacked generalization,''
  \emph{Journal of Machine Learning Research}, vol.~6, pp. 1751--1782, Dec
  2005.

\bibitem{ozay_icip}
M.~Ozay and F.~T.~Y. Vural, ``A new decision fusion technique for image
  classification,'' in \emph{Proceedings of the 16th {IEEE} the International
  Conference on Image Processing, ({ICIP} 2009)}, Cairo, Egypt, Nov 2009, pp.
  2189--2192.

\bibitem{Dzeroski}
S.~D\v{z}eroski and B.~\v{Z}enko, ``Is combining classifiers with stacking
  better than selecting the best one?'' \emph{Machine Learning}, vol.~54,
  no.~3, pp. 255--273, Mar 2004.

\bibitem{wolf}
L.~Wolf, T.~Hassner, and Y.~Taigman, ``Effective unconstrained face recognition
  by combining multiple descriptors and learned background statistics,''
  \emph{IEEE Transactions on Pattern Analysis and Machine Intelligence},
  vol.~33, no.~10, pp. 1978--1990, 2011.

\bibitem{wolf2}
O.~Kliper-Gross, T.~Hassner, and L.~Wolf, ``The action similarity labeling
  challenge,'' \emph{IEEE Transactions on Pattern Analysis and Machine
  Intelligence}, vol.~34, no.~3, pp. 615--621, 2012.

\bibitem{Zhouyu}
Z.~Fu, G.~Lu, K.~M. Ting, and D.~Zhang, ``A survey of audio-based music
  classification and annotation,'' \emph{IEEE Transactions on Multimedia},
  vol.~13, no.~2, pp. 303--319, 2011.

\bibitem{Cho}
S.-B. Cho and J.~H. Kim, ``Multiple network fusion using fuzzy logic,''
  \emph{IEEE Transactions on Neural Networks}, vol.~6, no.~2, pp. 497--501, Mar
  1995.

\bibitem{fvn}
L.~I. Kuncheva, ``"fuzzy" versus "nonfuzzy" in combining classifiers designed
  by boosting,'' \emph{IEEE Transactions on Fuzzy Systems}, vol.~11, no.~6, pp.
  729--741, Dec 2003.

\bibitem{11}
S.~K. Pal and S.~Mitra, ``Multilayer perceptron, fuzzy sets, and
  classification,'' \emph{IEEE Transactions on Neural Networks}, vol.~3, no.~5,
  pp. 683--697, Sep 1992.

\bibitem{12}
K.~E. Graves and R.~Nagarajah, ``Uncertainty estimation using fuzzy measures
  for multiclass classification,'' \emph{IEEE Transactions on Neural Networks},
  vol.~18, no.~1, pp. 128--140, Jan 2007.

\bibitem{short}
R.~D.~S. II and K.~Fukunaga, ``The optimal distance measure for nearest
  neighbor classification,'' \emph{IEEE Transactions on Information Theory},
  vol.~27, no.~5, pp. 622--626, 1981.

\bibitem{march}
E.~Marchiori, ``Class conditional nearest neighbor for large margin instance
  selection,'' \emph{IEEE Transactions on Pattern Analysis and Machine
  Intelligence}, vol.~32, no.~2, pp. 364--370, Feb 2010.

\bibitem{march2}
------, ``Hit miss networks with applications to instance selection,'' \emph{J.
  Mach. Learn. Res.}, vol.~9, pp. 997--1017, Jun 2008.

\bibitem{li}
Y.~Li and L.~Maguire, ``Selecting critical patterns based on local geometrical
  and statistical information,'' \emph{IEEE Transactions on Pattern Analysis
  and Machine Intelligence}, vol.~33, no.~6, pp. 1189--1201, Jun 2011.

\bibitem{garc}
S.~Garcia, J.~Derrac, J.~Cano, and F.~Herrera, ``Prototype selection for
  nearest neighbor classification: Taxonomy and empirical study,'' \emph{IEEE
  Transactions on Pattern Analysis and Machine Intelligence}, vol.~34, no.~3,
  pp. 417--435, Mar 2012.

\bibitem{fw}
F.~Fernandez and P.~Isasi, ``Local feature weighting in nearest prototype
  classification,'' \emph{IEEE Transactions on Neural Networks}, vol.~19,
  no.~1, pp. 40--53, 2008.

\bibitem{derrac}
J.~Derrac, I.~Triguero, S.~Garcia, and F.~Herrera, ``Integrating instance
  selection, instance weighting, and feature weighting for nearest neighbor
  classifiers by coevolutionary algorithms,'' \emph{IEEE Transactions on
  Systems, Man, and Cybernetics, Part B: Cybernetics}, vol.~42, no.~5, pp.
  1383--1397, 2012.

\bibitem{nca}
J.~Goldberger, S.~Roweis, G.~Hinton, and R.~Salakhutdinov, ``Neighbourhood
  components analysis,'' in \emph{Advances in Neural Information Processing
  Systems 17}, L.~K. Saul, Y.~Weiss, and L.~Bottou, Eds.\hskip 1em plus 0.5em
  minus 0.4em\relax Cambridge, MA: MIT Press, 2005, pp. 513--520.

\bibitem{lmnn}
K.~Q. Weinberger and L.~K. Saul, ``Distance metric learning for large margin
  nearest neighbor classification,'' \emph{J. Mach. Learn. Res.}, vol.~10, pp.
  207--244, Jun 2009.

\bibitem{Shalev}
S.~Shalev-Shwartz, Y.~Singer, and A.~Y. Ng, ``Online and batch learning of
  pseudo-metrics,'' in \emph{Proceedings of the Twenty First International
  Conference on Machine learning}, ser. ICML '04.\hskip 1em plus 0.5em minus
  0.4em\relax New York, NY, USA: ACM, 2004, pp. 94--101.

\bibitem{Paredes}
R.~Paredes and E.~Vidal, ``Learning weighted metrics to minimize
  nearest-neighbor classification error,'' \emph{IEEE Transactions on Pattern
  Analysis and Machine Intelligence}, vol.~28, no.~7, pp. 1100--1110, 2006.

\bibitem{duda}
R.~Duda, P.~Hart, and D.~Stork, \emph{Pattern Classification}.\hskip 1em plus
  0.5em minus 0.4em\relax New York, NY, USA: Wiley, 2001.

\bibitem{cover}
T.~Cover and P.~Hart, ``Nearest neighbor pattern classification,'' \emph{IEEE
  Transactions on Information Theory}, vol.~13, no.~1, pp. 21--27, Jan 1967.

\bibitem{fknn}
J.~Keller, M.~Gray, and J.~Givens, ``A fuzzy k-nearest neighbor algorithm,''
  \emph{IEEE Transactions on System, Man, and Cybernetics}, vol. SMC-15, no.~4,
  pp. 580 --585, 1985.

\bibitem{comb_kun}
L.~I. Kuncheva, \emph{Combining Pattern Classifiers: Methods and Algorithms},
  1st~ed.\hskip 1em plus 0.5em minus 0.4em\relax Haboken, NJ, USA:
  Wiley-Interscience, 2004.

\bibitem{hastie}
T.~Hastie and R.~Tibshirani, ``Discriminant adaptive nearest neighbor
  classification,'' \emph{IEEE Transactions on Pattern Analysis and Machine
  Intelligence}, vol.~18, no.~6, pp. 607--616, Jun 1996.

\bibitem{w}
R.-E. Fan, P.-H. Chen, and C.-J. Lin, ``Working set selection using second
  order information for training support vector machines,'' \emph{Journal of
  Machine Learning Research}, vol.~6, pp. 1889--1918, Dec 2005.

\bibitem{blake}
\BIBentryALTinterwordspacing
C.~B.~D. Newman and C.~Merz, ``{UCI} repository of machine learning
  databases,'' 1998. [Online]. Available:
  \url{http://www.ics.uci.edu/$\sim$mlearn/MLRepository.html}
\BIBentrySTDinterwordspacing

\bibitem{caltech}
P.~Gehler and S.~Nowozin, ``On feature combination for multiclass object
  classification,'' in \emph{IEEE 12th International Conference on Computer
  Vision}.\hskip 1em plus 0.5em minus 0.4em\relax IEEE, 2009, pp. 221--228.

\bibitem{gpu_knn}
V.~Garcia, E.~Debreuve, F.~Nielsen, and M.~Barlaud, ``k-nearest neighbor
  search: fast {GPU}-based implementations and application to high-dimensional
  feature matching,'' in \emph{IEEE International Conference on Image
  Processing (ICIP)}, Hong Kong, China, September 2010.

\bibitem{devroye}
L.~Devroye, L.~Gy{\"{o}}rfi, and G.~Lugosi, \emph{A Probabilistic Theory of
  Pattern Recognition}.\hskip 1em plus 0.5em minus 0.4em\relax Springer, 1996.

\bibitem{mpeg1}
H.~Eidenberger, ``\BIBforeignlanguage{English}{Statistical analysis of
  content-based mpeg-7 descriptors for image retrieval},''
  \emph{\BIBforeignlanguage{English}{Multimedia Systems}}, vol.~10, no.~2, pp.
  84--97, 2004.

\bibitem{mpeg2}
P.~Salembier and T.~Sikora, \emph{Introduction to MPEG-7: Multimedia Content
  Description Interface}, B.~Manjunath, Ed.\hskip 1em plus 0.5em minus
  0.4em\relax New York, NY, USA: John Wiley \& Sons, Inc., 2002.

\bibitem{mfcc}
O.~Lartillot and P.~Toiviainen, ``A matlab toolbox for musical feature
  extraction from audio,'' in \emph{Proceedings of the 10th International
  Conference on Digital Audio Effects}, Bordeaux, France, Sep 2007, pp.
  237--244.

\bibitem{hist_ent}
K.~F. Wallis, ``A note on the calculation of entropy from histograms,''
  Department of Economics, University of Warwick, UK, Tech. Rep., 2006.

\end{thebibliography}
\bibliographystyle{IEEEtran}

\end{document}